\documentclass[letterpaper, 10 pt, conference]{ieeeconf}  % Comment this line out if you need a4paper

\IEEEoverridecommandlockouts                              % This command is only needed if 
\usepackage{graphics} % for pdf, bitmapped graphics files
\usepackage{subcaption}
\usepackage{epsfig} % for postscript graphics files
\usepackage{amsmath} % assumes amsmath package installed
\usepackage{amssymb}  % assumes amsmath package installed

\title{\LARGE \bf
Learning Motor Resonance in Human-Human and Human-Robot Interaction with Coupled Dynamical Systems}

\author{Nuno Ferreira Duarte$^{1}$, Mirko Rakovi\'{c}$^{1,2}$, and 
        Jos\'{e} Santos-Victor$^{1}$
\thanks{This work was EU H2020 project 752611 - ACTICIPATE, FCT project UID/EEA/50009/2013 and RBCog-Lab research infrastructure.}
\thanks{$^{1}$Nuno Ferreira Duarte, Mirko Rakovi\'{c}, and Jos\'{e} Santos-Victor are with Vislab, Institute for Systems and Robotics, Instituto Superior T\'{e}cnico, Universidade de Lisboa, Portugal 
        {\tt\small$\{$nferreiraduarte, rakovicm, jasv$\}$@isr.tecnico.ulisboa.pt}}%
\thanks{$^{2}$M Rakovi\'{c} is with Faculty of Technical Sciences, University of Novi Sad, Novi Sad, Serbia
        {\tt\small rakovicm@uns.ac.rs}}%
}

\begin{document}

\maketitle
\thispagestyle{empty}
\pagestyle{empty}

%%%%%%%%%%%%%%%%%%%%%%%%%%%%%%%%%%%%%%%%%%%%%%%%%%%%%%%%%%%%%%%%%%%%%%%%%%%%%%%%
\begin{abstract}
Human interaction involves very sophisticated non-verbal communication skills like understanding the goals and actions of others and coordinating our own actions accordingly. Neuroscience refers to this mechanism as motor resonance, in the sense that the perception of another person’s actions and sensory experiences activates the observer's brain as if (s)he would be performing the same actions and having the same experiences.

We analyze and model non-verbal cues (arm movements) exchanged between two humans that interact and execute handover actions. The contributions of this paper are the following: (i) computational models, using recorded motion data, describing the motor behaviour of each actor in action-in-interaction situations; (ii) a computational model that captures the behaviour of the `´giver'' and ``receiver'' during an object handover action, by coupling the arm motion of both actors; and (iii) embedded these models in the iCub robot for both action execution and recognition.

Our results show that: (i) the robot can interpret the human arm motion and recognize handover actions; and (ii) behave in a ``human-like'' manner to receive the object of the recognized handover action.

\end{abstract}

\section{INTRODUCTION} \label{sec:intro}

Humans interact with each other in all types of environments, e.g. at work or at home. These interactions happen very frequently and may involve physical objects present in the surrounding space. In specific scenarios, humans can seamlessly perform a sequence of tasks which require multiple agents (people) to interact with each other in order to reach a common goal, \cite{SEBANZ2006JointAction}. 

A core element in interaction situations is the need and the means of expressing an intent. Intent can be communicated directly by comprehensive vocalized sentence or encoded as non-verbal cues through body, head, or eye movements. For instance, the mirror neuron system found in humans and other primates, may have the fundamental function of enabling the preparation of an appropriate complementary response to an observed action. It may explain how two individuals can become so attuned to cooperating in joint actions \cite{Rozzi2015Grasping}.
    \begin{figure}[t]
      \centering
      \includegraphics[width = 0.48\textwidth]{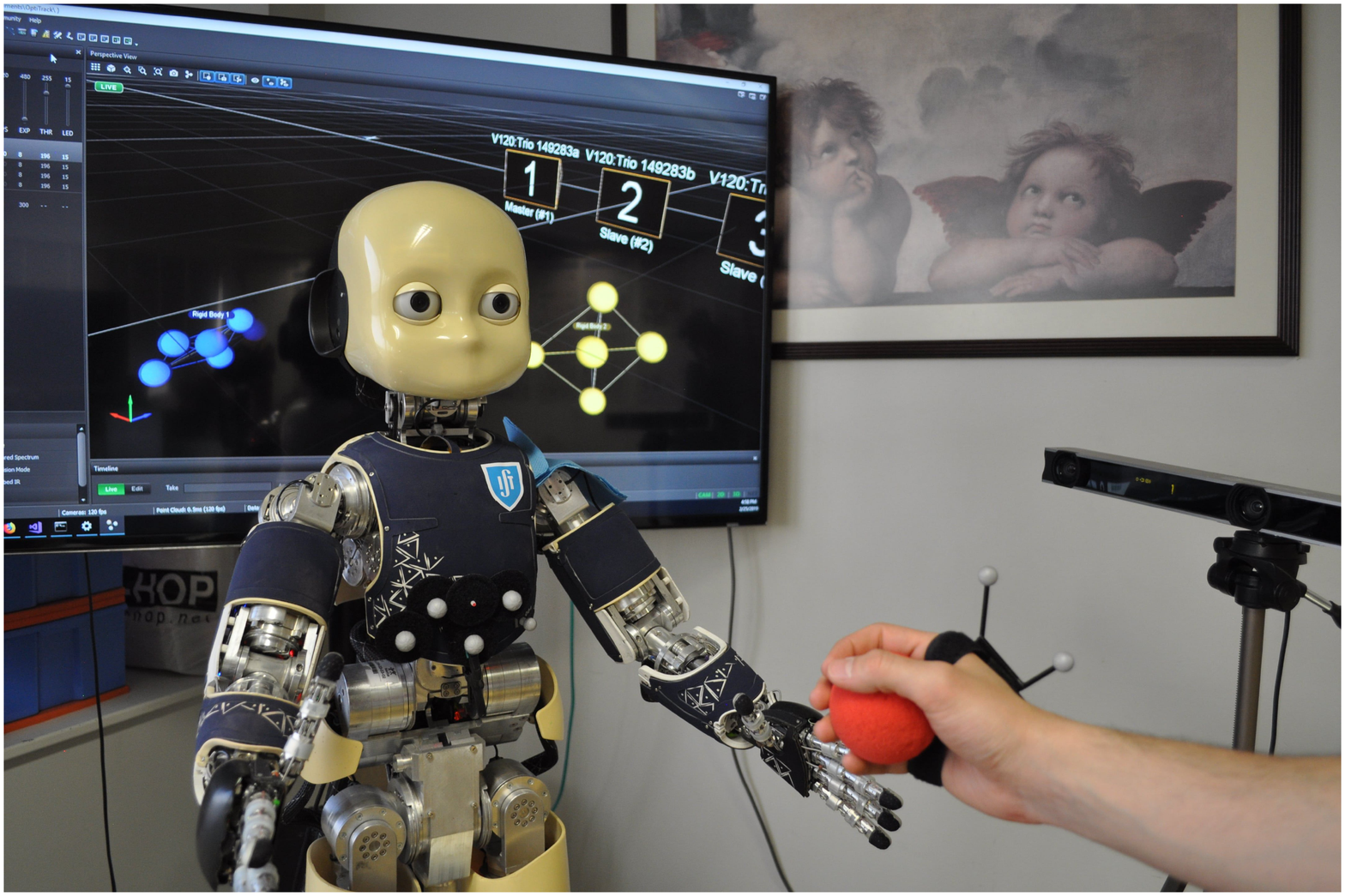}
      \caption{Experimental setup: the human is the leader of the action and the movement of the arm is tracked by the markers on the wrist using the OptiTrack bar. The iCub robot acts as the follower of the action and the adapts to the behavior of the human.} 
      \label{fig:hriExp}
	\end{figure}

When human collaborators are required to perform actions which involve sharing objects between each other, coordination and understanding are pivotal factors in a successful interaction. Research in corticomuscular and intermuscular coherence in humans report the advantages of non-verbal communication in such interactions. Synchronisation in motor coordination \cite{ibarra2018synchronization} is seen as a biological condition in order to improve efficiency and reliability in human-human interaction (HHI). Moreover, synchronisation between two agents is preferred, to two individuals systems, to achieve optimal motor control. Further research on psychology \cite{nowak2017functional}, cognition, and neuroscience \cite{huyi2017brain-brain}, reinforces on the idea that social interaction adheres from synchronisation of lower-level elements \cite{bassetti2017action-in-interaction}. 

For robotics, a logical step is for people to connect to robots, and build robotic platforms that use non-verbal communication to express intent, and understand human intention. 

With this long-term research goal in mind, we start by analyzing the HHI scenario, in order to understand the non-verbal communication cues between humans. The experimental scenario used is the same as in \cite{rakovic2018dataset} where dyadic interaction in a turn-taking game involving actions-in-interaction. We extracted and recorded the wrist location of both participants throughout the whole interaction. 

The collected data were then used to model the behaviour of both participants in every dyad, for action-in-interaction tasks such as handing over an object. The computational framework uses time-invariant Dynamical Systems (DS), and it generates two models corresponding to person performing the handing over (leader) and the other person receiving the object (follower). We established a link between the two dynamical systems using Coupled Dynamical System (CDS), which combines the dynamics of performing a handover action with the dynamics of being on the receiving-end. 
The ability to establish this coupling or synchronization between the movements of two interaction parties, is the cornerstone of the full power of non-verbal communication during interaction.
%Encapsulating the coupling, a.k.a the synchronisation, between two agent's motor movement we established an action-in-interaction.
The developed coupled model was incorporated in the iCub humanoid robot, for validating the model in a HRI scenario, and to test the robustness to external perturbations in real time and in a real scenario. 

In the final sections of the paper, we discuss the results obtained from our HRI experiments, as well as the correlation with the human-human synchronization in the HHI experiments. Our results demonstrate that it is possible to model the coupling between two humans in an action-in-interaction situation, and transfer that model into a controller that elicits the same interaction behavior between a human and a humanoid robot. The last section is reserved for the conclusions and work planned for the future.

\section{RELATED WORK-} \label{sec:soa}

The field of HRI has been the focus of many research efforts during the last few decades. A key observation is that, successful human-robot collaborations requires that the human has to understand the robot and, simultaneously, that the robot is also able to understand the human actions \cite{sciutti2018humanizing}. This has propelled fields such as robotics, computer vision, human cognition, psychology, and neuroscience, to combine resources in the aim of proposing techniques that tackle those intertwined issues. 

Lukic et al. \cite{lukic2014learning} have conducted human behavioral studies, with the added knowledge of cognitive neuroscience, to present an intrapersonal motor control model for manipulating objects inspired in the coordination of the human hand, arm, and eyes movements. This model is able to successively couple the hand-arm-eyes movements, based on the  the visuomotor coordination of eyes-arm-hand movements, that human's demonstrate when executing tasks of obstacle avoidance and manipulation. This coupling is achieved using CDS to link the eyes-arm and arm-hand controllers. The coupling function is a combination of Gaussian Mixture Models which, in turn, can also generate the human-like behavior/movement coordination. Following that principle, \cite{duarte2018action} studied the human behavior in a dyadic interaction where the movement of the eyes and arm was analyzed to model the repertoire of non-verbal communication cues, when one agent interacts with others. This model was then adapted to HRI to replicate the human-like behavior in a robot. The experiment validated the intention understanding of robot's intention was at a very high level. Further extension is mutual understanding of the behavior of both agents in a dyadic interaction. \cite{mirrazavi2018unified} also use DS for multiple robotic arms in order to coordinate to reach a moving object. However, none of the aforementioned work involve coordinating motion for a dyadic interaction. 

Rakovi\'{c} et al. \cite{rakovic2018gazedialogue} models the gaze behavior interdependencies for two agents using Hidden Markov Models (HMM) where the states are the gaze fixations of one agent, and the observations are the gaze fixations of the other. HMM then estimates the valid gaze behavior of the agent observing the action, and the correct action performed by the other agent. This is used to build a controller which permits the robot (observer/follower) to read the non-verbal communication cues from the gaze of the human participant (leader of the action) and infer the intended action. On another note, \cite{marin2009interpersonal} state that to improve social interactions, robots and humans should mutually influence each other. As such, \cite{duarte2018actionalignment} worked on action alignment between humans and robots. Where the robot is the leader, and in order to correctly adapt to the intricacies of collaborating with a human, a Discrete Time Markov Model is used to give information whether the human is understanding the action of the robot.

The work discussed above has focused on either intrapersonal coupling, synchronization, and action understanding from the gaze behavior of two agents. Notwithstanding, understanding of the arm movements in humans and robots is also a focus of extensive work. Dragan et al. \cite{dragan2013legibility} have created a framework for evaluating predictable and legible arm movements for humans and robots. Where legibility is am exaggerated motion towards the end-goal, and the predictable is a natural human movement. Nonetheless, there is no guarantee of generalization for all types of scenarios.  yet no understanding whether this is present in HHI, and in retrospect, HRI scenarios. Although, to improve social interactions, robots and humans should mutually influence each other \cite{marin2009interpersonal}. 

One approach proposed for an efficient human-robot collaboration is to predict the human movement during the experiment and optimize the path for the robot to avoid colliding with the human \cite{sisbot2010synthesizing, ding2011human}. Although, this may work for some scenarios, it does not solve the core problem of HRI, the mutual understanding of each one's action. For these cases, only the robot is understanding the action of the human. One solution found is to delay the handover task of the robot, since that has proven to improve the human understanding of the robot's intention  \cite{admoni2014deliberate}. 

Synchronization between humans is present in interactive behaviours such as walking together, talking, object manipulation, etc \cite{mortl2014rhythm}. \cite{andry2011using, hasnain2012synchrony} have began analysing the importance of synchrony in humans and the applications to human-robot scenarios. Robots that look more human-like, the participants tended to synchronize better their movements. Hence, correlation between synchronization and appearance of the robot gives the conclusion that humanoid robots have an advantage when it comes to interacting with humans. 

\section{METHODOLOGY} \label{sec:metho}

This section contains the formalism of Dynamical Systems (DS) and the estimation of its parameters with a Gaussian Mixture Model (GMM). Then, we introduce the formalism of Coupled Dynamical Systems (CDS) which entails learning a coupling motion between two DS. We extend the motor coordination from the body parts of a single individual, as presented in [9], to the movement coordination across two individuals engaged in a scenario of action-in-interaction assignments.

\subsection{Dynamical Systems (DS)}

Let $\pmb{\xi}(t) \in \mathbb{R}^{d}$ denote the state vector. Considering $\textit{N}$ demonstrations of the interaction, we define $\{{{\pmb{\xi}}^{t}}_n, {{\dot{\pmb{\xi}}}^{t}}_n \}$, $\forall t \in [0, T_n]$, $n \in [1, N]$, where ${{\pmb{\xi}}^{t}}_n$ and  ${{\dot{\pmb{\xi}}}^{t}}_n$ are respective the state vector and its derivative, evaluated at time $t$ for the $n$-th demonstration. The number of samples in the $n$-th action is represented by $T_n$. The collected data are instances of motions which can be represented as first-order differential equations:
\begin{equation}
\pmb{\dot{\xi}} = \pmb{\textnormal{f}}(\pmb{\xi}) + \in
\end{equation}
where $\pmb{\textnormal{f}}:\mathbb{R}^{d} \to \mathbb{R}^{d}$ is a continuous and continuously differential function, with a single equilibrium point $\pmb{\dot{\xi}}^* = \pmb{\textnormal{f}}(\pmb{\xi}^*) + \in$. The zero mean Gaussian noise $\in$ allows it to handle errors such as motion variability. For spatial perturbations, the $\pmb{\xi}$ is set in the reference frame of the target. The DS is encoded using GMM which defines a joint distribution function $\mathcal{P}({{\pmb{\xi}}^{t}}_n, {\dot{\pmb{\xi}^{t}}}_n)$ over the collected data as mixture of $K$ Gaussian distributions \cite{khansari2011learning}. 

\subsection{Coupled Dynamical Systems (CDS)}

Using CDS enables the integration of two independent DS, learning the coupling function between them \cite{shukla2012coupled}. This coupling behaviour takes inspiration from the biological studies on motor synchronisation of reach-grasp coupling \cite{mitz1991learning}, 
% I COULD HAVE ADDED ANOTHER (7 FROM SHUKLA %
as well as the coupling between humans \cite{mortl2014rhythm}. CDS consist of a master-slave system, where the master sub-system is the DS of the first motor dynamics $\mathcal{P}({{\xi}^{t}}_n, {{\dot{\xi}}^{t}}_n | {\theta^{g}}_{master})$. After encoding the master, the next step is to infer the state of the slave conditioned on the master, $\mathcal{P}(\Psi({{\xi}^{t}}_n), {{\xi}^{t}}_n | {\theta^{g}}_{coupled})$. This is the coupling which then allows to encode the dynamics of the slave sub-system $\mathcal{P}({{\xi}^{t}}_n, {{\dot{\xi}}^{t}}_n | {\theta^{g}}_{slave})$, $\forall g \in \mathcal{G}$. $\Psi:\mathbb{R}^{d_m} \to \mathbb{R}$ is the coupling function which is a function dependent on the master state. 

Lukic et. al \cite{lukic2014learning} extended the method to couple the hand grip aperture, the arm motion, and saccadic eye movements during object manipulation and obstacle avoidance. They composed a CDS with three DS, corresponding to the dynamics of the eyes, arm, and hand, integrated with two coupling functions, one to infer the position of the arm concerning the movement of the eyes, and another to infer the finger configuration of the hand to the location of the arm. 

\subsection{Extended CDS for Human-Human Coordination} 
\label{sec:extend}

One of our core contributions is to extend the motor coupling modeling and control to action-in-interaction scenarios, linking  the body movements of two humans during handover actions. We argue that each agent has its own internal CDS architecture of \cite{lukic2014learning} in order to manipulate objects and avoid obstacles. While interacting with other agents, an external CDS architecture administers the motor communication and coordination required for a successful interaction. In a dyadic scenario where two agents participate in an action-in-interaction task, the first agent (the leader) performs the action, while the second agent (the follower) observes and reacts to the leader's actions. Each agent's CDS is an internal model defined by the intrapersonal coordination of \cite{lukic2014learning}. The end-effector of the follower agent is then conditioned on the end-effector of the leading agent:

\begin{equation} \label{ds:agent1}
\mathcal{P}({{\xi}}_{ef_1}, {{\dot{\xi}}}_{ef_1} | {\theta^{g}}_{1}) 
\end{equation} 
\begin{equation} \label{ds:coupling}
\mathcal{P}(\Psi({{\xi}}_{ef_1}), {{\xi}}_{ef_2} | {\theta^{g}}_{1:2})
\end{equation} 
\begin{equation} \label{ds:agent2}
\mathcal{P}({{\xi}}_{ef_2}, {{\dot{\xi}}}_{ef_2} | {\theta^{g}}_{2})
\end{equation} 
where (\ref{ds:agent1}) represents the furthest DS (the end-effector of the intrapersonal coordination) of the CDS of agent 1, which encodes the dynamics of the end-effector and generates the out most state of agent 1. This state is given to the coupling sub-system in (\ref{ds:coupling}) and applied to the coupling function $\Psi({\xi}_{ef_1})$ in order to infer the state of agent 2, which in return is used to encode the dynamics using (\ref{ds:agent2}).

The primary reason for the use of dynamical systems to model the intrapersonal motor coordination as well as the agent-to-agent motor coordination is its stability to input errors. It seamlessly permits a robot to conform its trajectory instantly in the face of perturbations, such as arm motion variability.

\section{Scenario} \label{sec:scenario}

The experimental scenario consists of a dyadic interaction between 2 participants where performing two types of actions: (i) \textit{placing} an object on a table, which falls into the category of \textit{individual} actions; and (ii) \textit{giving} an object to the other participant, that belongs to the category of \textit{actions-in-interaction}. The experiment is explained in greater detail along with the procedure for data collection, and what type of sensory data, in the paper associated to the dataset \cite{rakovic2018dataset}. 

The experiment was arranged in such a manner that neither participant could perceive the intention of the other participant. This allows for a natural movement by humans during the interaction.  The Cartesian coordinates of the wrist of each participant's right arm, used in the experiment, was recorded using the OptiTrack motion capture system. A total of 72 right-hand wrist trajectories were recorded of a handover action-in-interaction: 36 corresponding to the leader of the action handing over the object, and the remainder 36 of the follower, who was receiving the object.

\section{MODELING} \label{sec:modeling}

This section explains the modeling of the arm movements of both agents using the approach described in Sectin~\ref{sec:extend}. Using the data described in Section~\ref{sec:scenario}, it is possible to define a coupling sub-system between the Cartesian Coordinates of the arm of agent 1 to compute the Cartesian Coordinates of agent 2. 

\subsection{Dynamics of each Agent}\label{sec:agent}

\begin{figure*}[htpb]
	\begin{subfigure}{0.24\textwidth}
	\includegraphics[trim={0cm 0 1cm 0},clip, width = 0.99\textwidth]{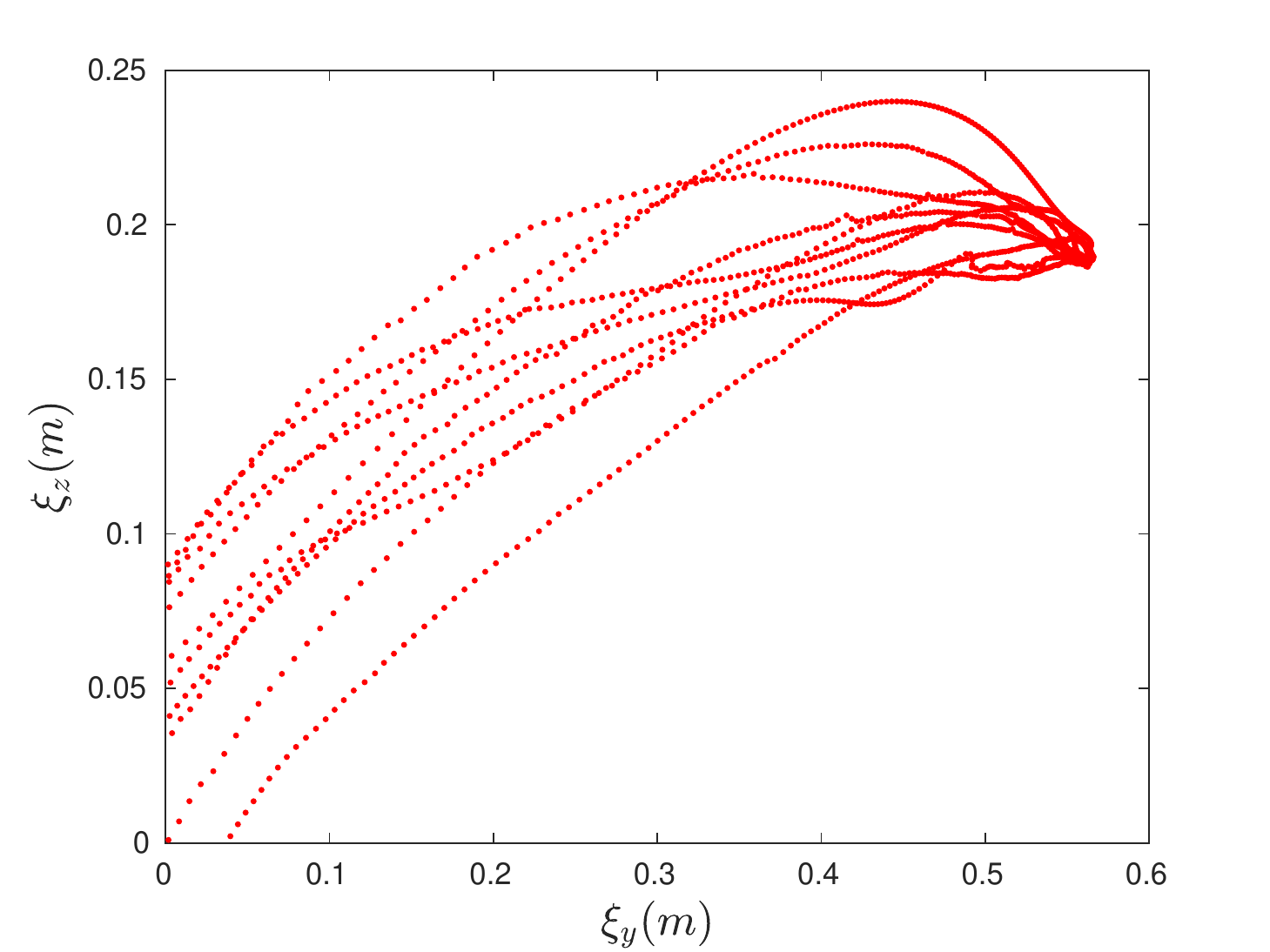}\label{subfig1:demo}
	\includegraphics[trim={0cm 0 1cm 0},clip, width = 0.99\textwidth]{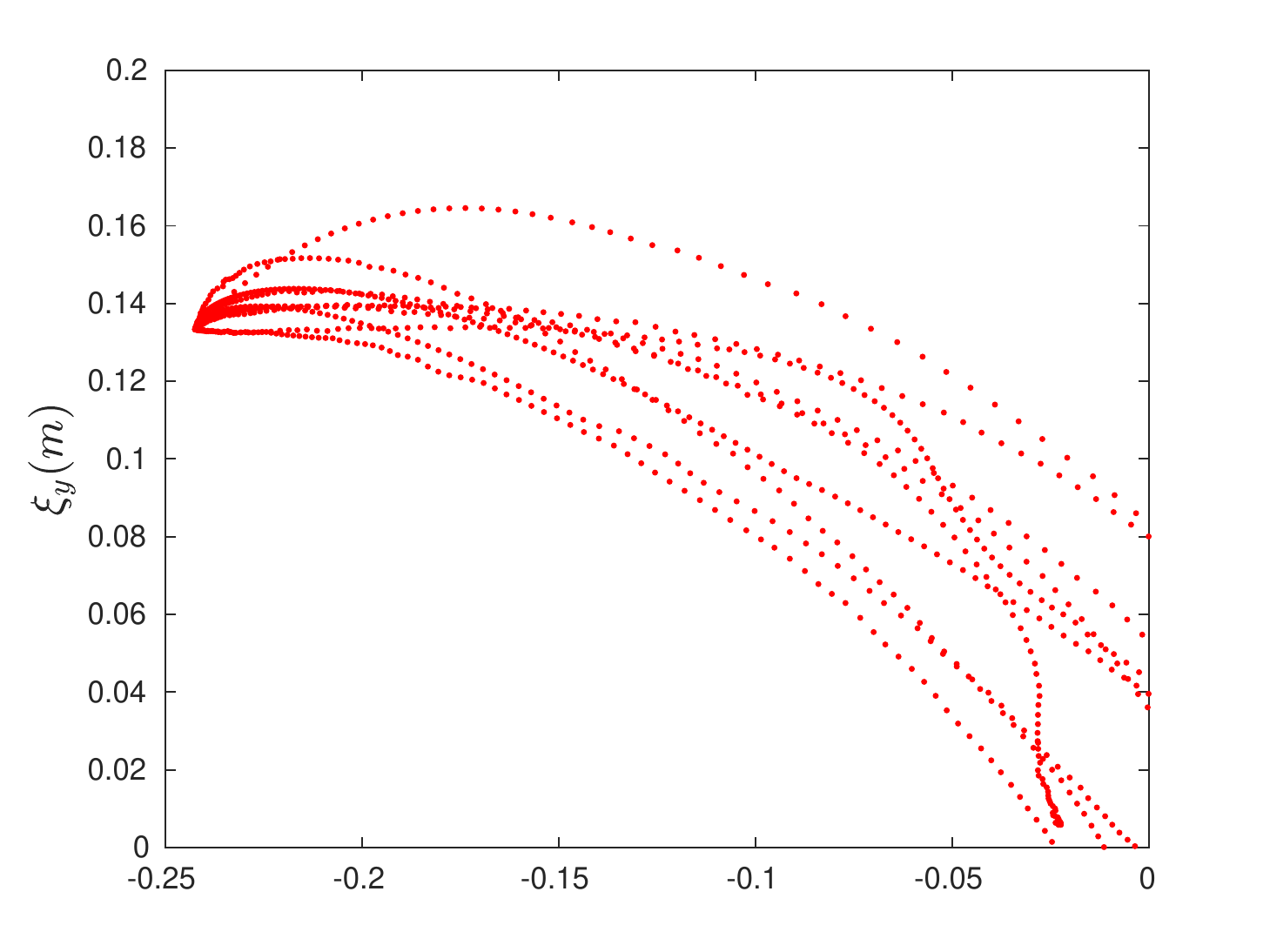}
    \caption{}\label{subfig2:demo}
    \end{subfigure}
	\begin{subfigure}{0.24\textwidth}
	\includegraphics[trim={0cm 0 1cm 0},clip, width = 0.99\textwidth]{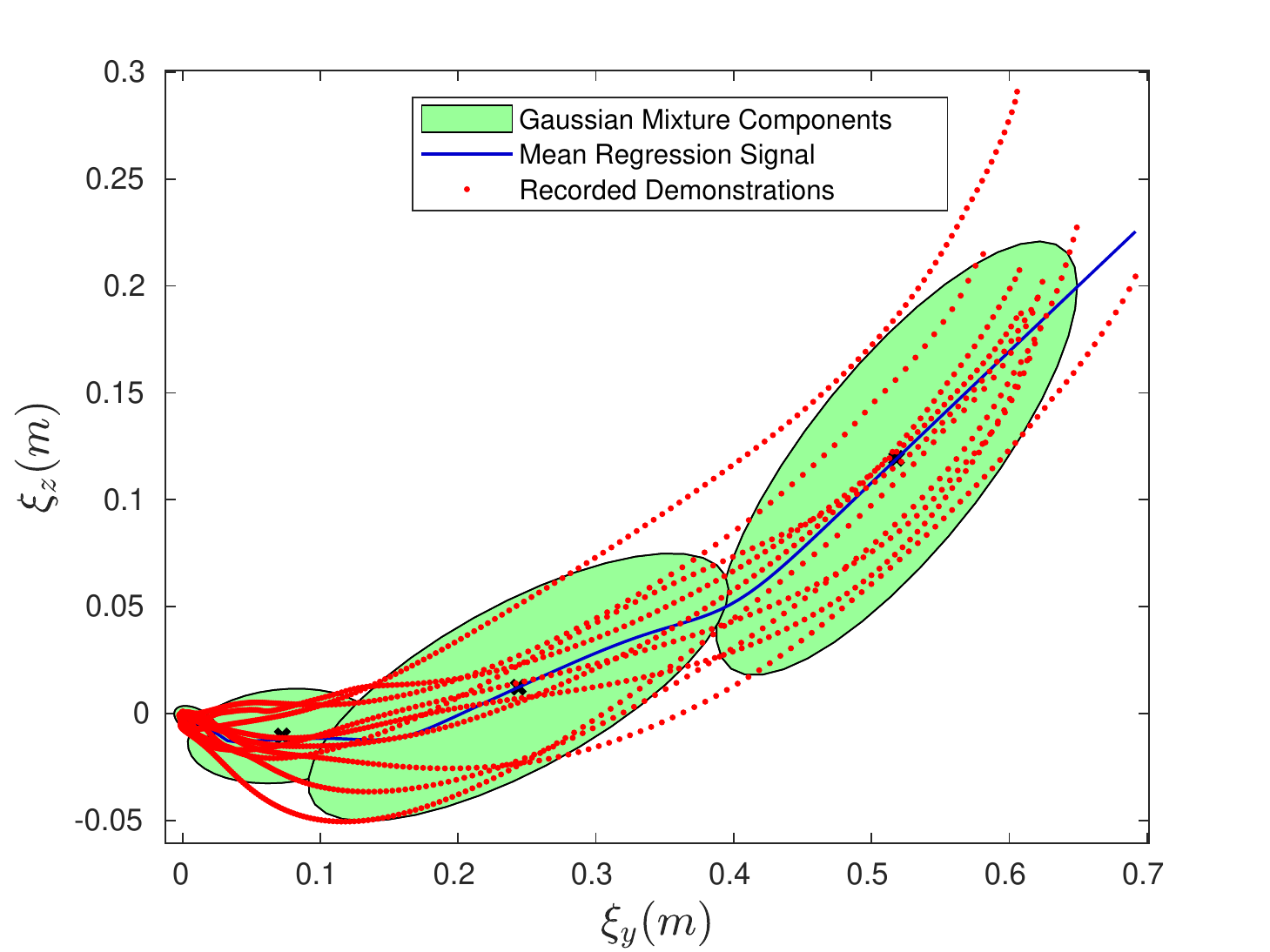}\label{subfig1:xy}
	\includegraphics[trim={0cm 0 1cm 0},clip, width = 0.99\textwidth]{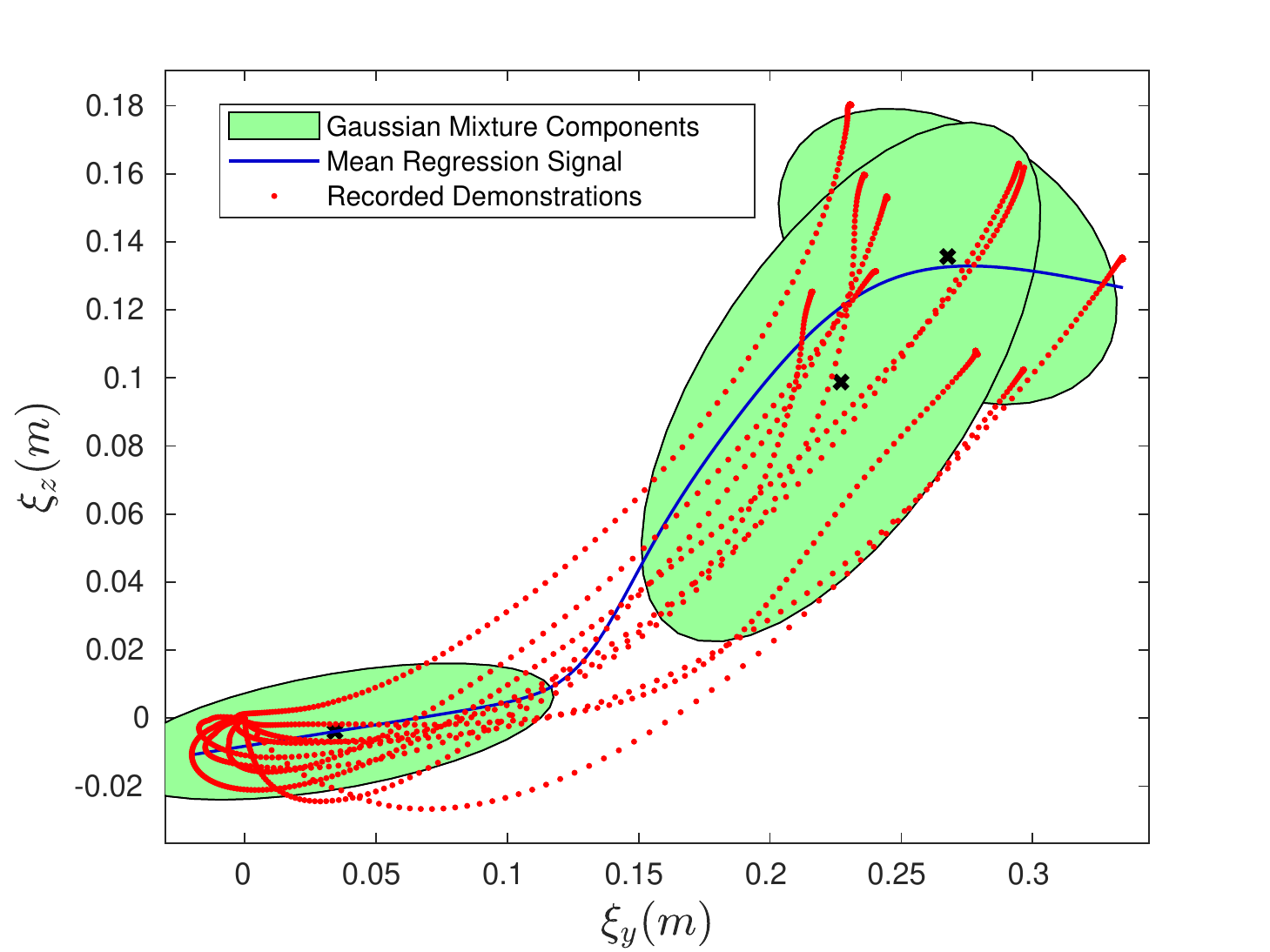}
    \caption{}\label{subfig2:xy}
    \end{subfigure}
    \begin{subfigure}{0.24\textwidth}
	\includegraphics[trim={0cm 0 1cm 0},clip, width = 0.99\textwidth]{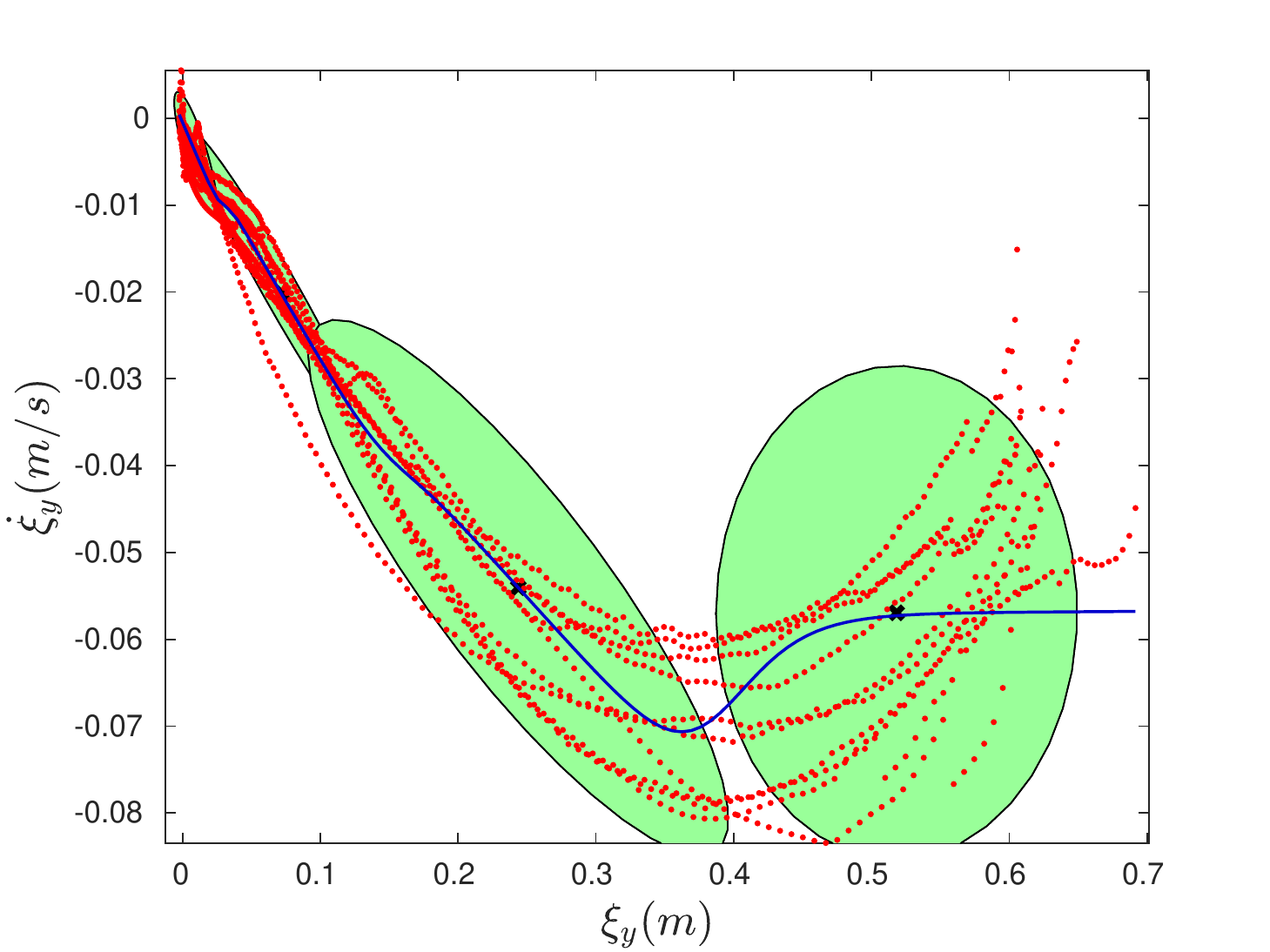}\label{subfig1:xdotx}    
	\includegraphics[trim={0cm 0 1cm 0},clip, width = 0.99\textwidth]{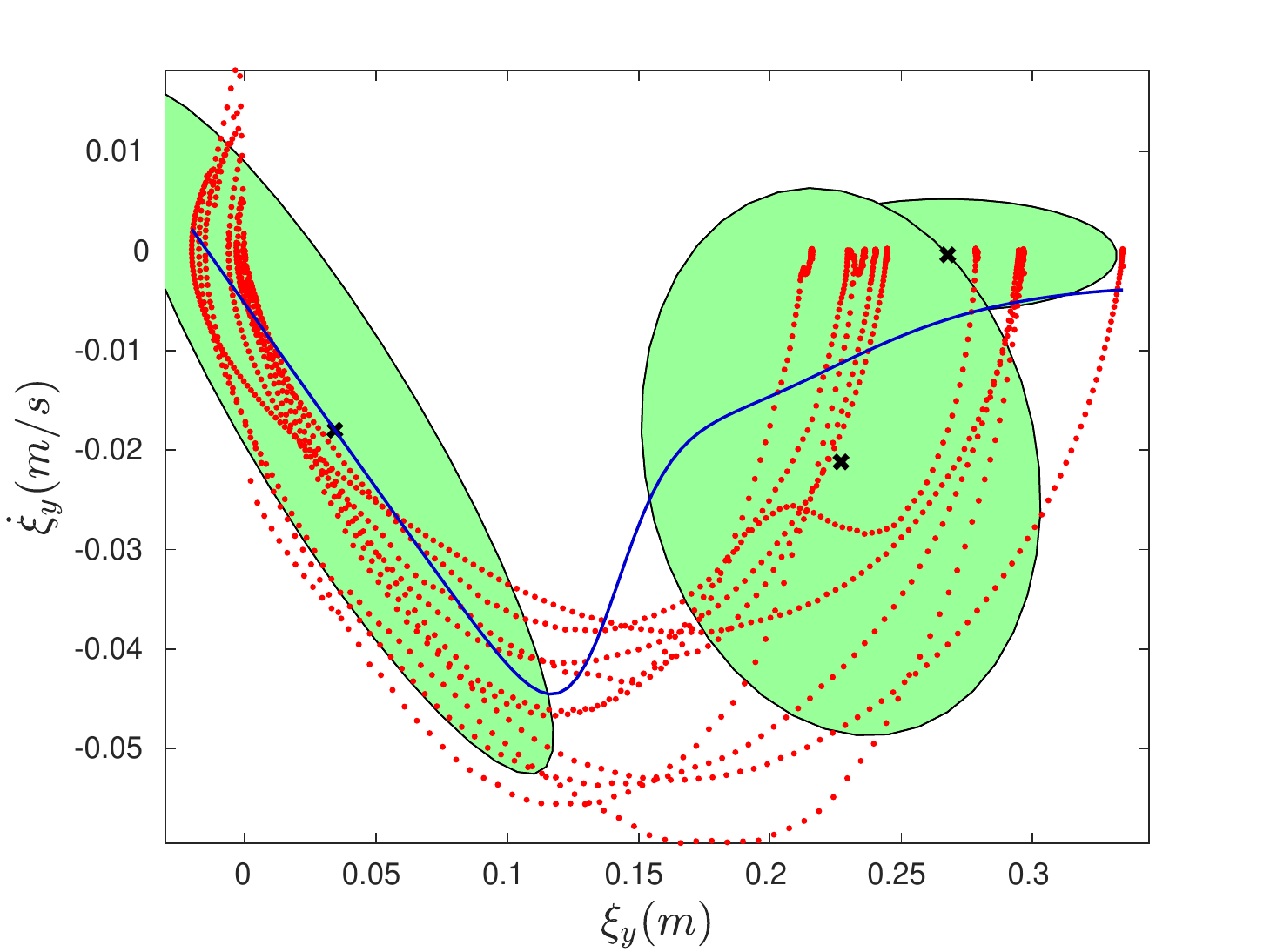}
    \caption{}\label{subfig2:xdotx}    
    \end{subfigure}
    \begin{subfigure}{0.24\textwidth}
	\includegraphics[trim={0cm 0 1cm 0},clip, width = 0.99\textwidth]{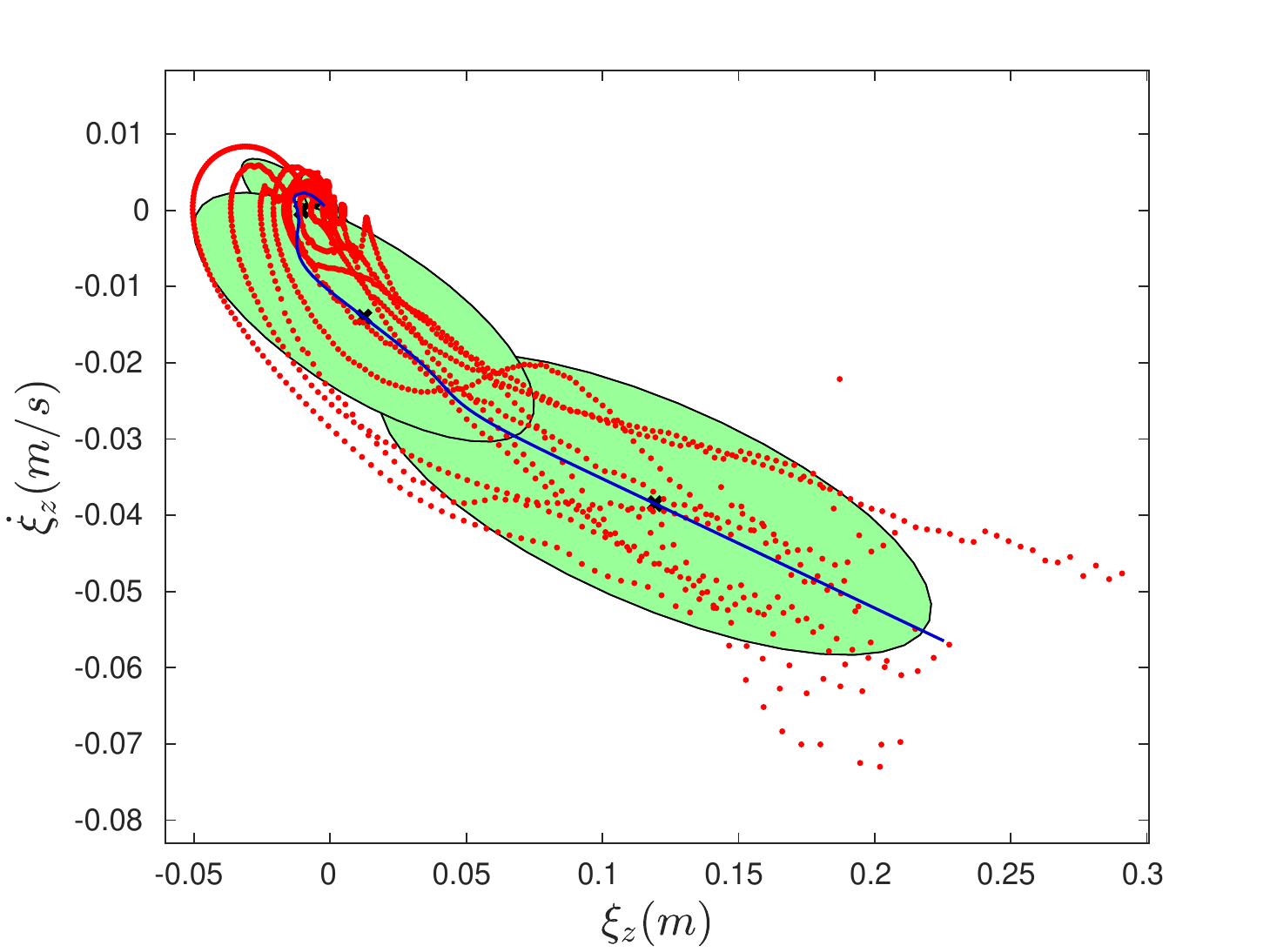}\label{subfig1:ydoty}   
	\includegraphics[trim={0cm 0 1cm 0},clip, width = 0.99\textwidth]{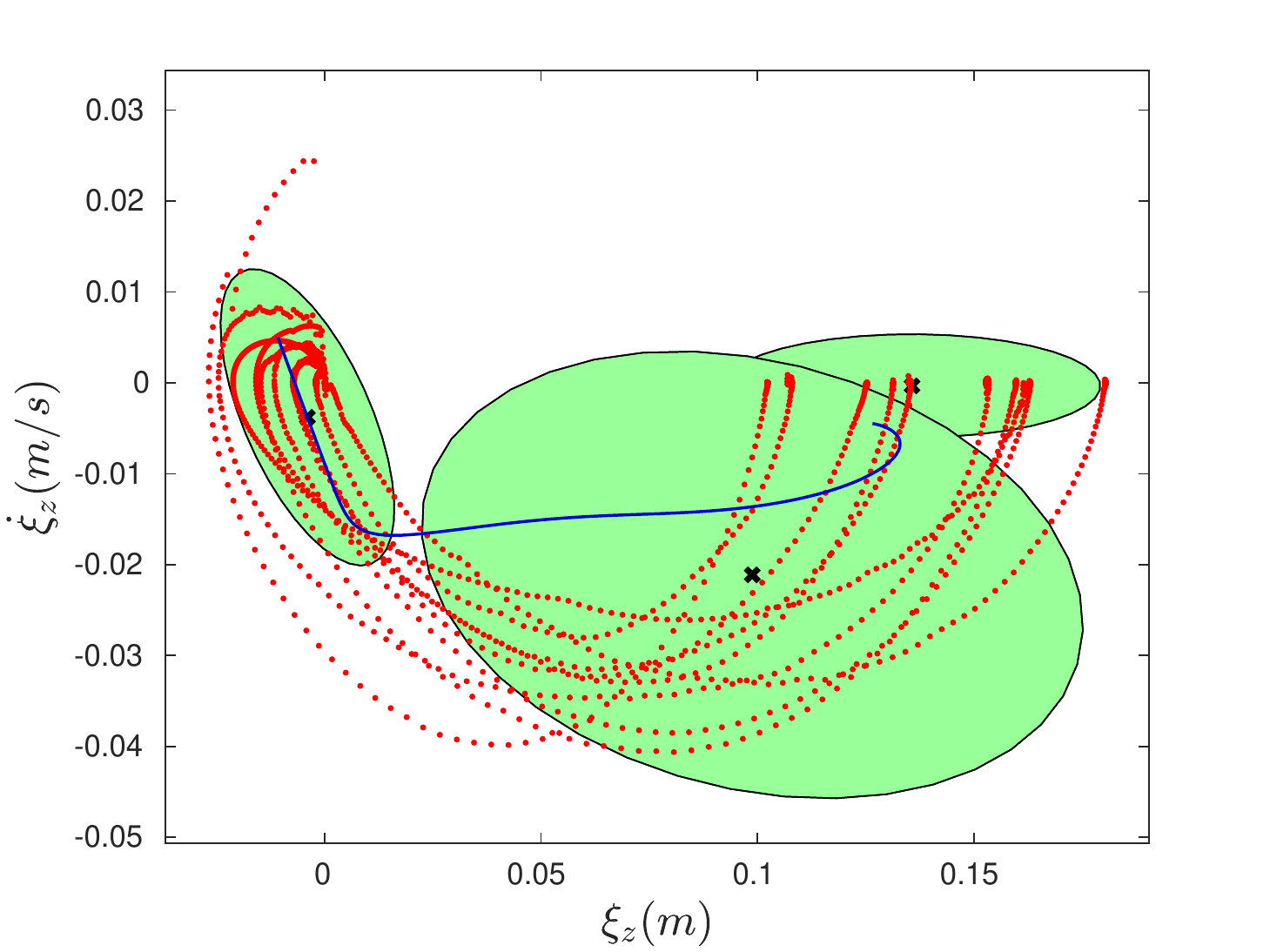}
    \caption{}\label{subfig2:ydoty}
    \end{subfigure}
\caption{Learned internal model for leader agent (top row), and follower (bottom row) - which due to our dataset only composing of right hand wrist, the intrapersonal model is defined as single DS of the arm dynamics. (a) is the recorded demonstrations for action-in-interaction; (b) is the GMM encoding the desired value of ${\xi}_{ef_z}$ (i.e. ${\hat{\xi}}_{ef_z}$) given the current value of ${\xi}_{ef_y}$; (c) the GMM encoding the velocity distribution conditioned on the y coordinate (proximity); (d) the GMM encoding the velocity distribution conditioned on the z coordinate (height). The Cartesian coordinates shown correspond to the major displacement in the action. }\label{fig:ds-agent1}
\end{figure*}

The arm movement of the leader perspective for the handover action is shown in Figure \ref{subfig2:demo}. In order to encode the dynamics of the arm movement the stability, i.e. the converging point, is set to be the handover point. To compute the GMM parameters we use the stable estimator of dynamical systems (SEDS) approach \cite{khansari2011learning}, since it ensures global stability in the individual GMMs. 

The GMM parameters are used to derived the Gaussian Mixture Regression (GMR) which provides the output data correspondent to the desired input values. In this case, from Figures \ref{subfig2:xdotx}, \ref{subfig2:ydoty}, it provides the desired velocity control for the particular position of the arm of agent 1. The next plots show the DS corresponding to the behavior of agent 2, a.k.a the follower in the interaction. 

Figure \ref{subfig2:demo} represents the corresponding human behavior to the handover. Figure \ref{subfig2:xy} shows the GMR mean trajectory generated from the GMM parameters of the follower's movement to the leader's action-in-interaction. Just as in the leader's action, the stability point of the dynamics of the follower's action is set as the handover point. 

Even though the wrist moves in three dimensions, we argue that, for the HHI scenario experiment, there were only two relevant dimensions to be considered. We use the Cartesian y-dimension coordinate to refer to the proximity of the wrist to the handover location, and z-dimension coordinate to refer to the height, perpendicular to the handover location. Moreover, we argue that the handover action can be generalized to any orientation. As long as the proximity and height constraints are fulfilled, the motion will be understood by humans, i.e. legible behaviour. 

\subsection{Coupling between Agents}
\label{sec:couple}

\begin{figure}[htpb]
	\begin{subfigure}{0.24\textwidth}
	\includegraphics[width = 0.99\textwidth]{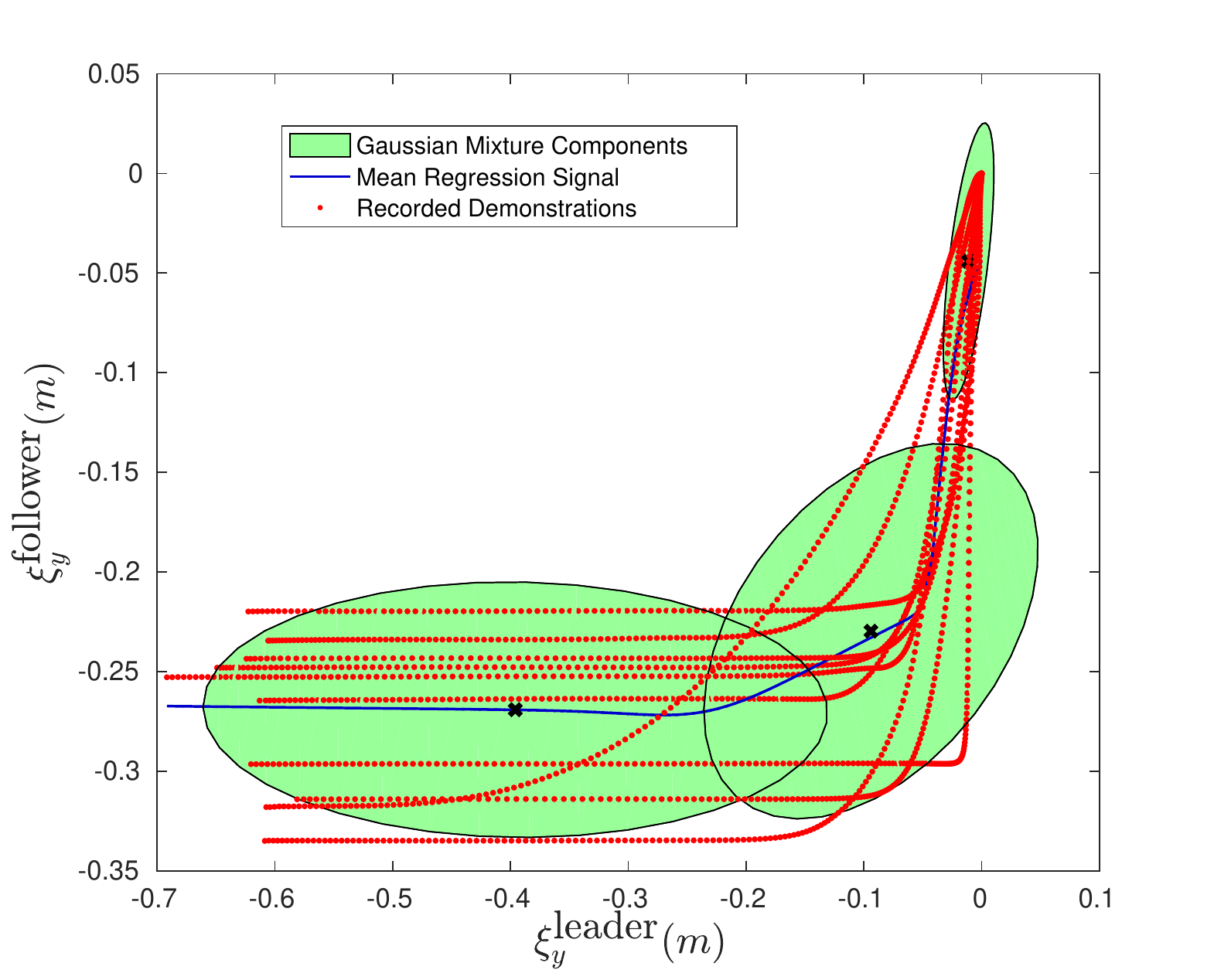}
    \caption{}\label{subfig:coupleY}
    \end{subfigure}
	\begin{subfigure}{0.24\textwidth}
	\includegraphics[width = 0.99\textwidth]{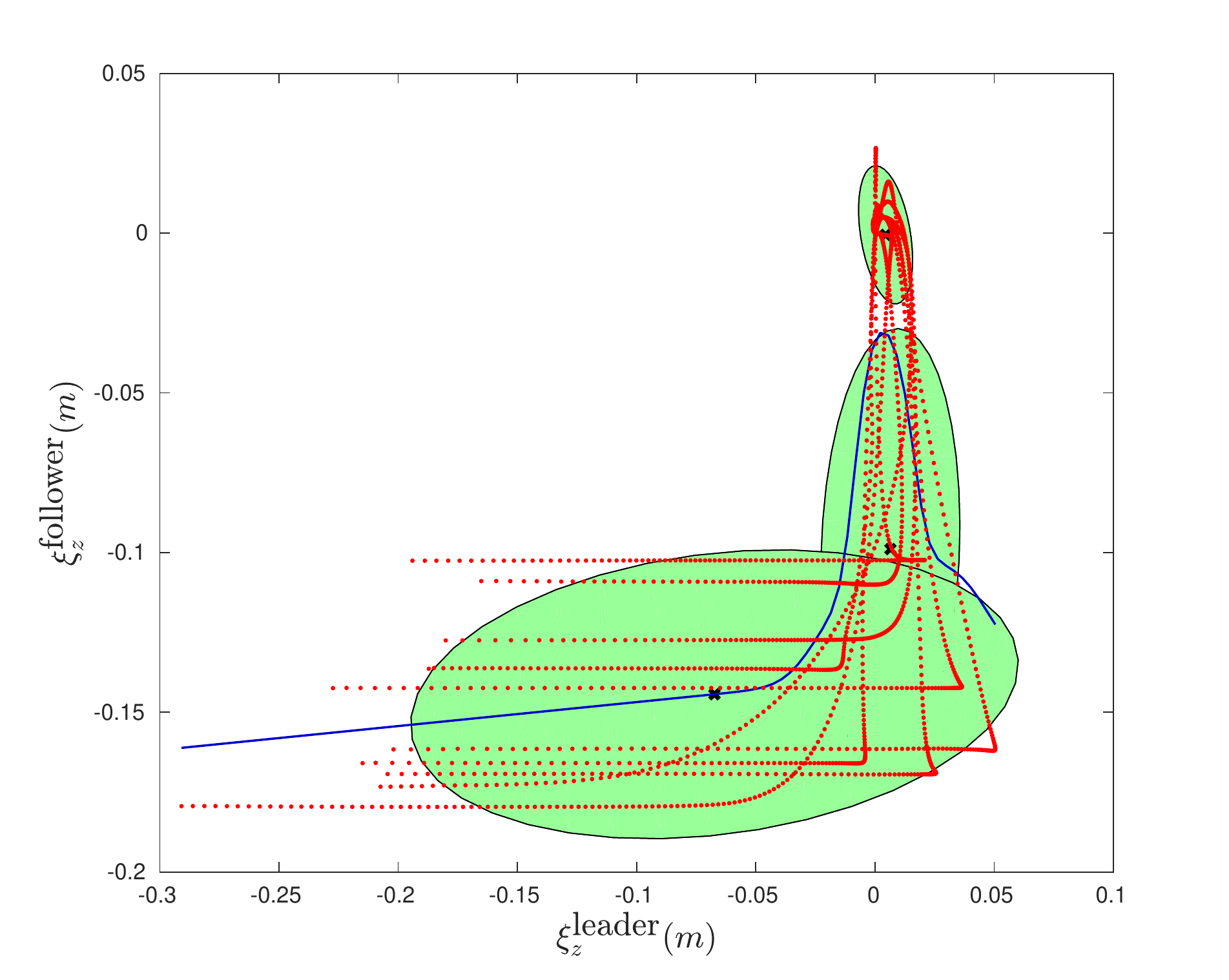}
    \caption{}\label{subfig:coupleZ}
    \end{subfigure}
    \caption{Learned CDS between agent 1 and agent 2 end-effectors: (a) coupling the proximity of the leader's wrist to the handover location, (b) coupling the elevation of the leader's wrist to the handover location.} \label{fig:couple}
\end{figure}

Regarding the agent-to-agent motor coordination, we chose to disambiguate the relevant Cartesian coordinates (y- and z-) during the handover action. Figure \ref{fig:couple} reveals the correlation between agent's wrist motion for the proximity and height coordinates, respectively. 

From the analysis of the coupling models the following can be concluded: 
The coupling function used is $\Psi({{\xi}}_{ef_1}) = \parallel y \parallel$, for the proximity, and $\Psi({{\xi}}_{ef_1}) = z$ for the height coordinate, and the values of $\alpha$, $\beta$ are to set to 1. Although the norm is considered to be biologically plausible metric, the height coordinate could not be normalized due to the variability of wrist motion when it comes to altitude. This is related to the initial start of the wrist which could vary the motion. The reason for coupling the arm-arm is bolstered by psychologists and neurobiology scientists as an indispensable factor for social interaction \cite{kelso2013outline,kendon1990conducting,feldman2007synchrony}. 

The GMMs that encode the dynamics between the arms of both agents are learned using the Expectation-Maximization (EM) algorithm \cite{nasrabadi2007pattern}. From analyzing the data of both arm movements side by side during the interaction, one observes an explicit synchronisation between the movement of the leader and follower arms during the handover action. This relation between the arm's of both participants is identified as the coupling between agent 1's non-verbal communication during the handover, and the corresponding non-verbal communication of agent 2 understanding. 

From this section, we can draw the following conclusions. Firstly, the influence of the leader's wrist behaviour upon the follower is stronger when close to the handover location. Secondly, the height coordinate has a more significant impact in closer distances than the proximity coordinate. Thirdly, for considerable distances (larger than 20 centimetres), the proximity coordinate already gives some inclination of the type of action. This might have to do with the setup of the HHI scenario. The scenario involved performing actions of handing over an object or placing them in a tower. The tower was located in between the handover location, which has created an ambiguity situation. Nonetheless, this gives certainty as to whether one motion of the arm is intended for a handover, as can be seen from the height coordinate. The coupling effect of the leader motion was only strongly present for altitudes close to the handover location. 

The overall conclusion that can be taken from the analysis is the existence of different coupling factors: (i) how the arm motion of one human can provide information about the action intent, and (ii) how the second human responds to that information and adapts according to what it infers. 

\section{HUMAN-ROBOT INTERACTION} \label{sec:hri}

In order to evaluate the models of the intrapersonal motor coordination and the coupling agent-to-agent motor coordination developed in Section \ref{sec:modeling}, we tested our humanoid robot in a HRI scenario, while using the controllers that embedded the biologically inspired arm motion models. 

\subsection{Experimental Setup}

\begin{figure}
	\includegraphics[width = 0.49\textwidth]{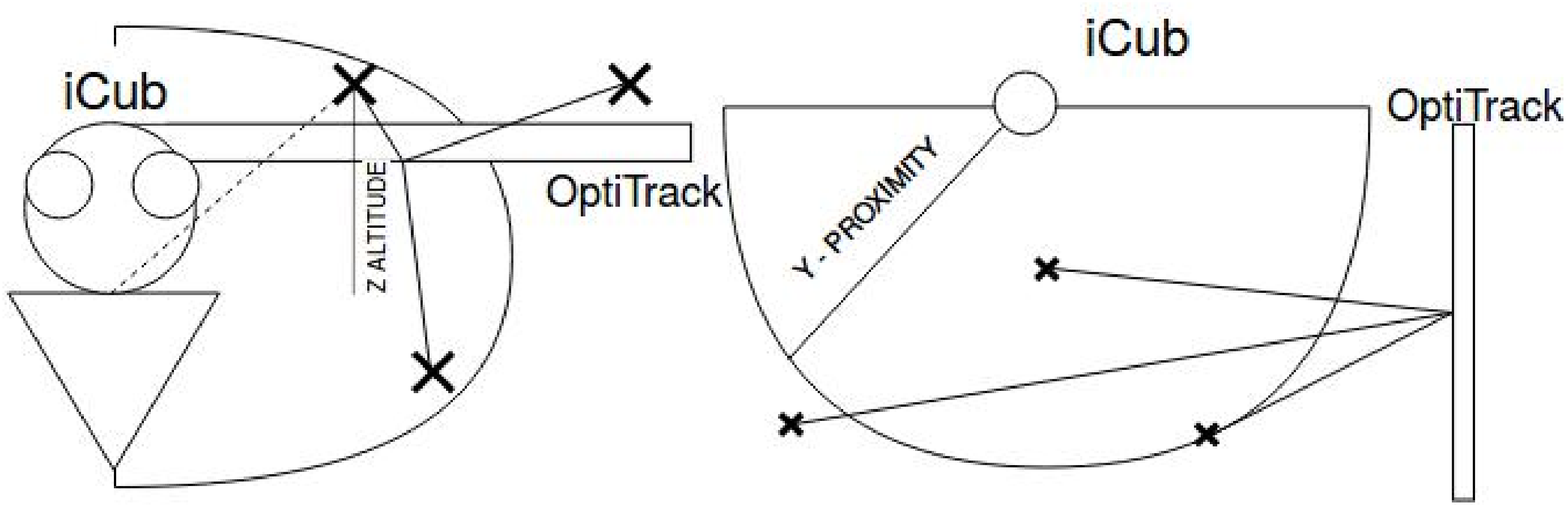}
    \caption{Illustration of HRI setup. Optitrack bar tracks the location of the human wrist and our controller computes the altitude, and proximity, with respect to the iCub center of mass. On the left side it represents the profile view of the experiment, the top view is illustrated on the right side. 3 example points are marked in the illustration and represented in both perspectives.}\label{fig:hri}
\end{figure}

The HRI scenario is shown in Figure \ref{fig:hri}. For each experiment, the human is the leader and performs a handover action. The action is intended for the iCub humanoid robot. The iCub is capable of performing actions that are legible to humans \cite{duarte2018action} as, to some extent, it possesses a similar motor. As such, it is ideal to analyze the arm movement generated by the robot controller. 

The human arm movements were recorded with an OptiTrack motion capture (MoCap) system. Markers were attached to (i) the iCub torso, (ii) wrist of the human, and rigid body movements were devised from them. As depicted in Figure \ref{fig:hri}, the first rigid body, in Cartesian coordinates frame, is positioned on the torso of the iCub, and the second rigid body is one of the represented crosses. 

The location of the human end-effector is expressed in Cartesian coordinates referenced at the iCub rigid body. The coordinates are sent from the Mocap system and streamed through a Lab Streaming Layer \cite{kothe2018lab} as a yarp port to the humanoid robotic controller. The coordinates are the input to the controller which then outputs the 3D Cartesian coordinates for the end-effector of the humanoid robot. 

\subsection{Validation}

\begin{figure}[htpb]
	\begin{subfigure}{0.24\textwidth}
	\includegraphics[width = 0.99\textwidth]{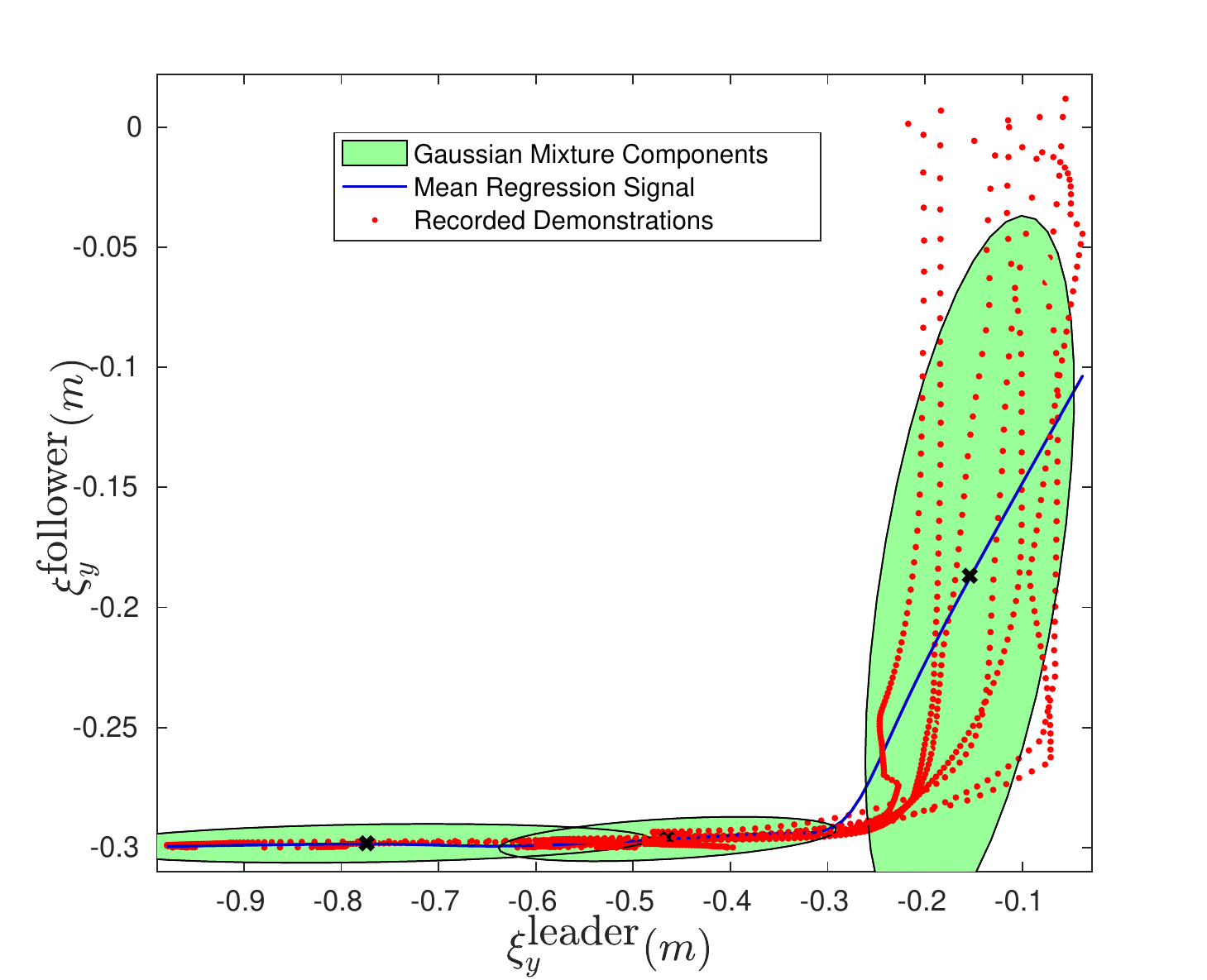}
    \caption{}\label{subfig:coupleYhri}
    \end{subfigure}
	\begin{subfigure}{0.24\textwidth}
	\includegraphics[width = 0.99\textwidth]{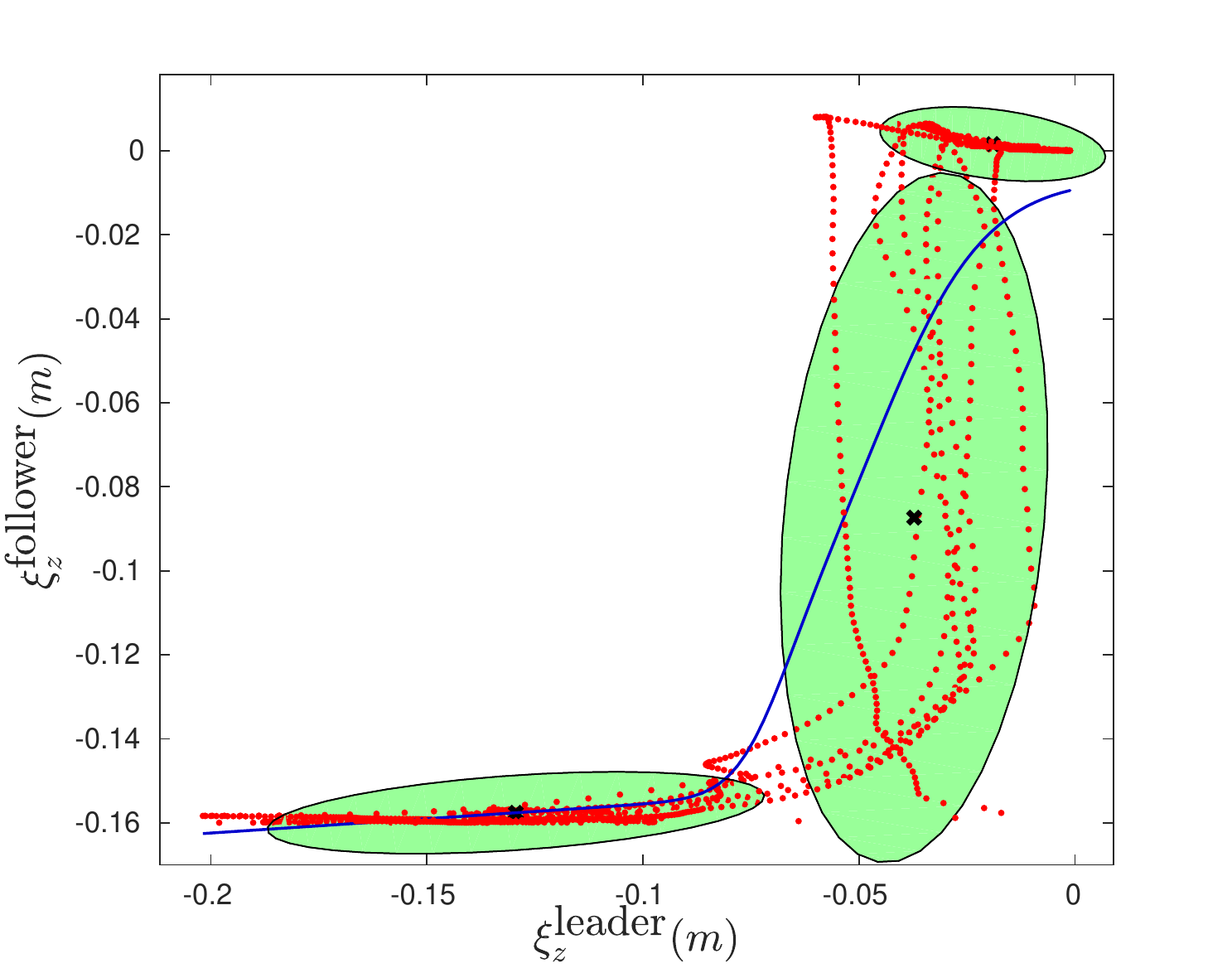}
    \caption{}\label{subfig:coupleZhri}
    \end{subfigure}
    \caption{Generated CDS from HRI experiments (iCubSIM vs Real Time Human wrist): (a) proximity, (b) height.} \label{fig:couplehri}
\end{figure}

From the HRI scenario, the behaviour of the iCub humanoid robot towards the human motion is analyzed. Figure \ref{fig:couplehri} shows the data of the human wrist collected with OptiTrack, and the corresponding output of humanoid robot from the model developed in Section \ref{sec:modeling}. From the data points we ran the EM algorithm to generate the dynamics between the human agent and the robot agent. The GMM parameters and the GMR trajectory represented in Figure \ref{fig:couplehri} reflect the coupling model observed in the HHI experiments. 

From the coupling model the corresponding coordinates for the robot are then processed with the internal DS of the follower to compute the desired velocity control, Figure \ref{subfig2:xdotx} and \ref{subfig2:ydoty}. The goal is to to control the wrist joint in a biologically inspired behaviour for the iCub humanoid robot. The trajectories of the motion performed by the robot support the human-like notion in \cite{duarte2018action}, indicating that humans can decode, from the motion of the arm, the intention of the action, either performed by a robot or other humans. 

\subsection{Results}

\begin{figure*}[htb]
    \centering \offinterlineskip
	\includegraphics[width = 0.115\textwidth]{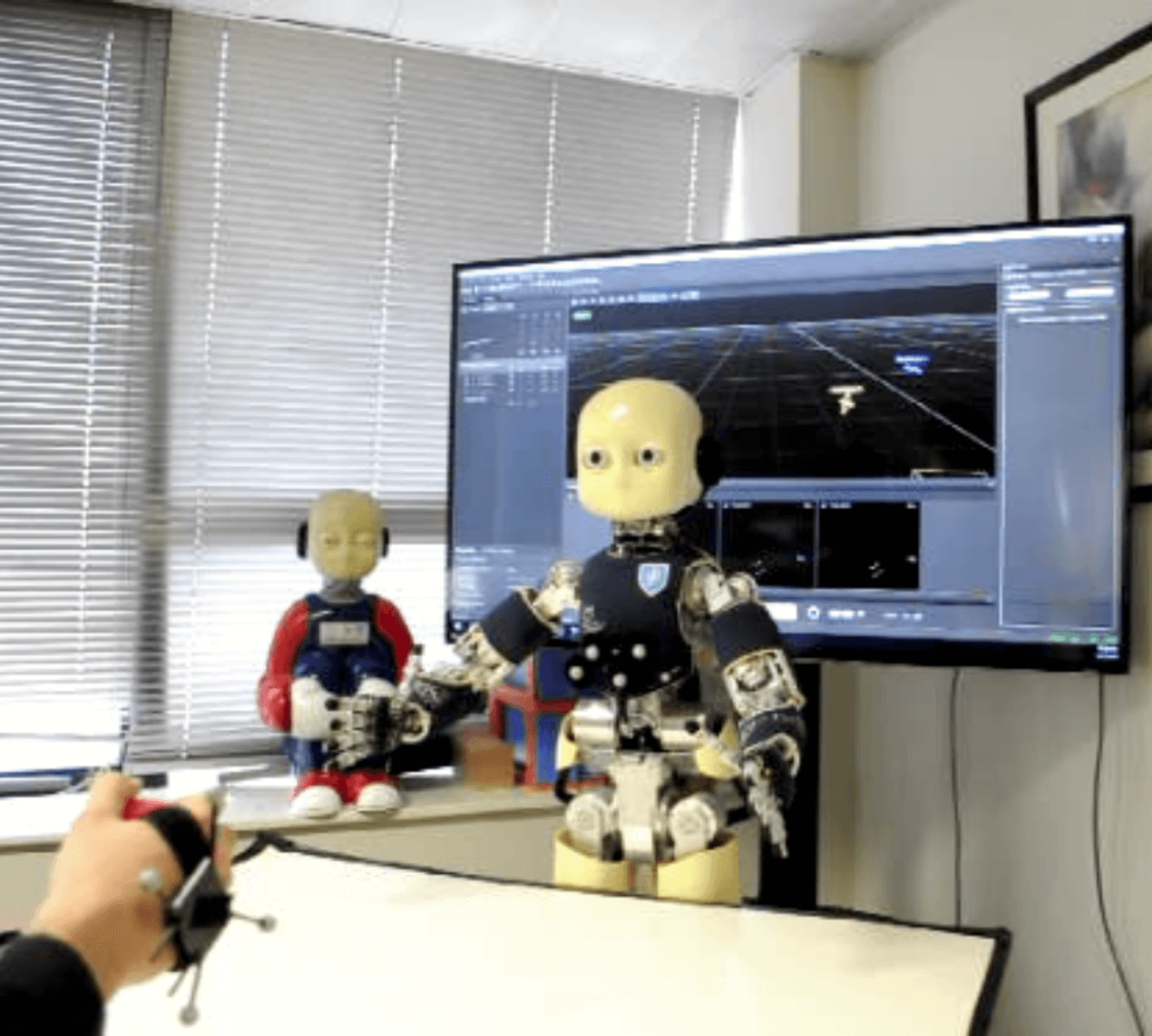}
	\includegraphics[width = 0.115\textwidth]{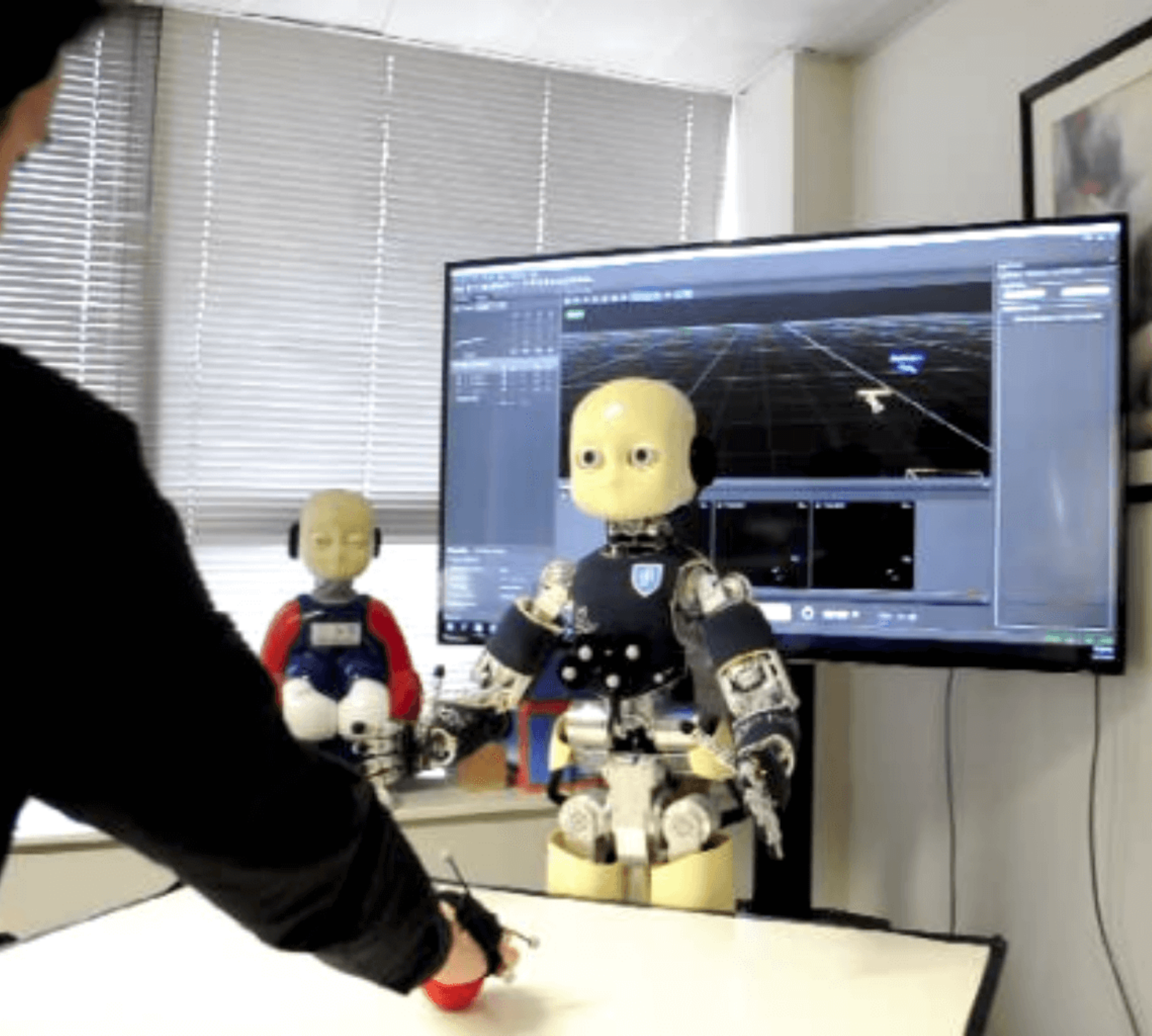}
	\includegraphics[width = 0.115\textwidth]{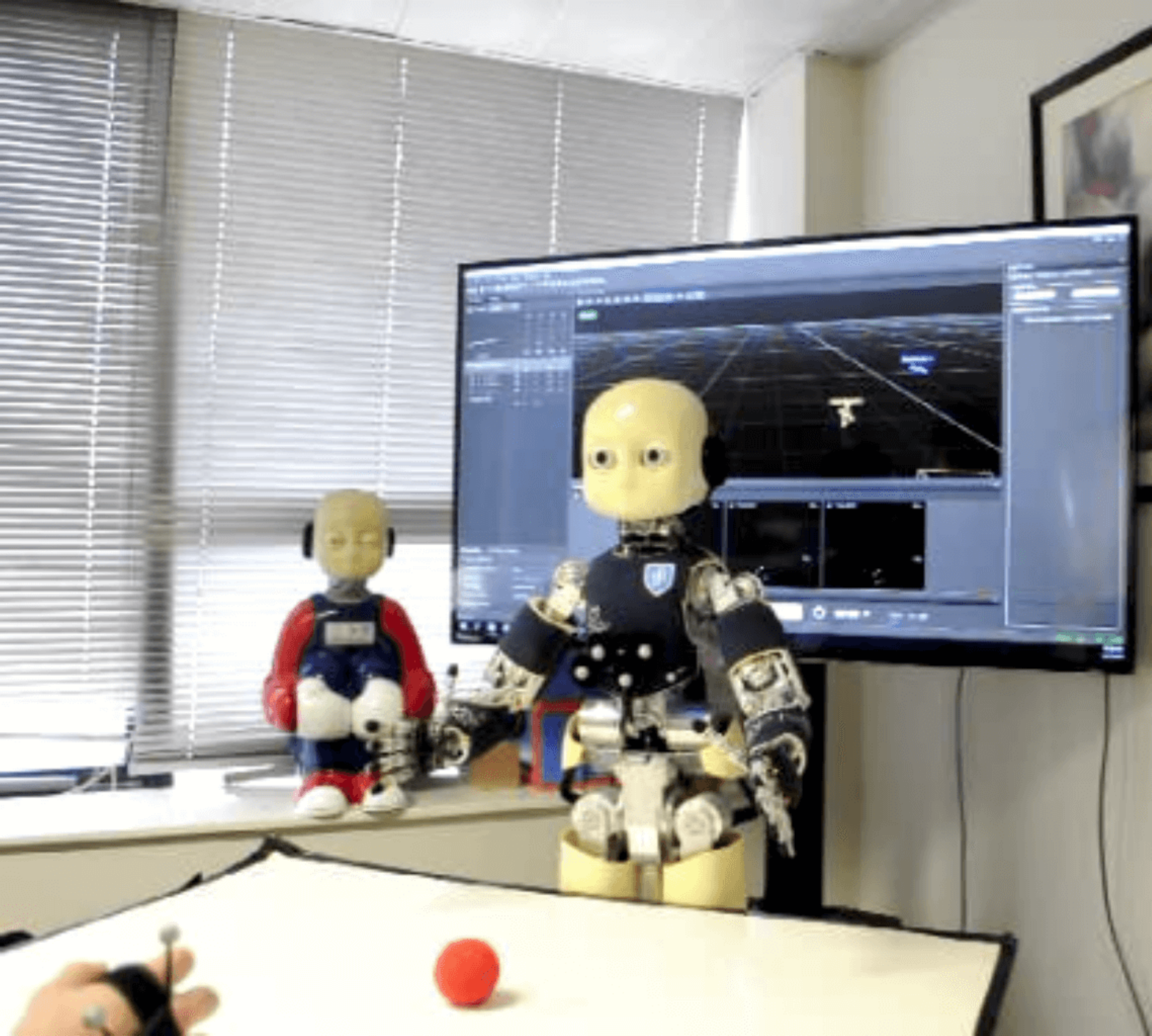}
	\includegraphics[width = 0.115\textwidth]{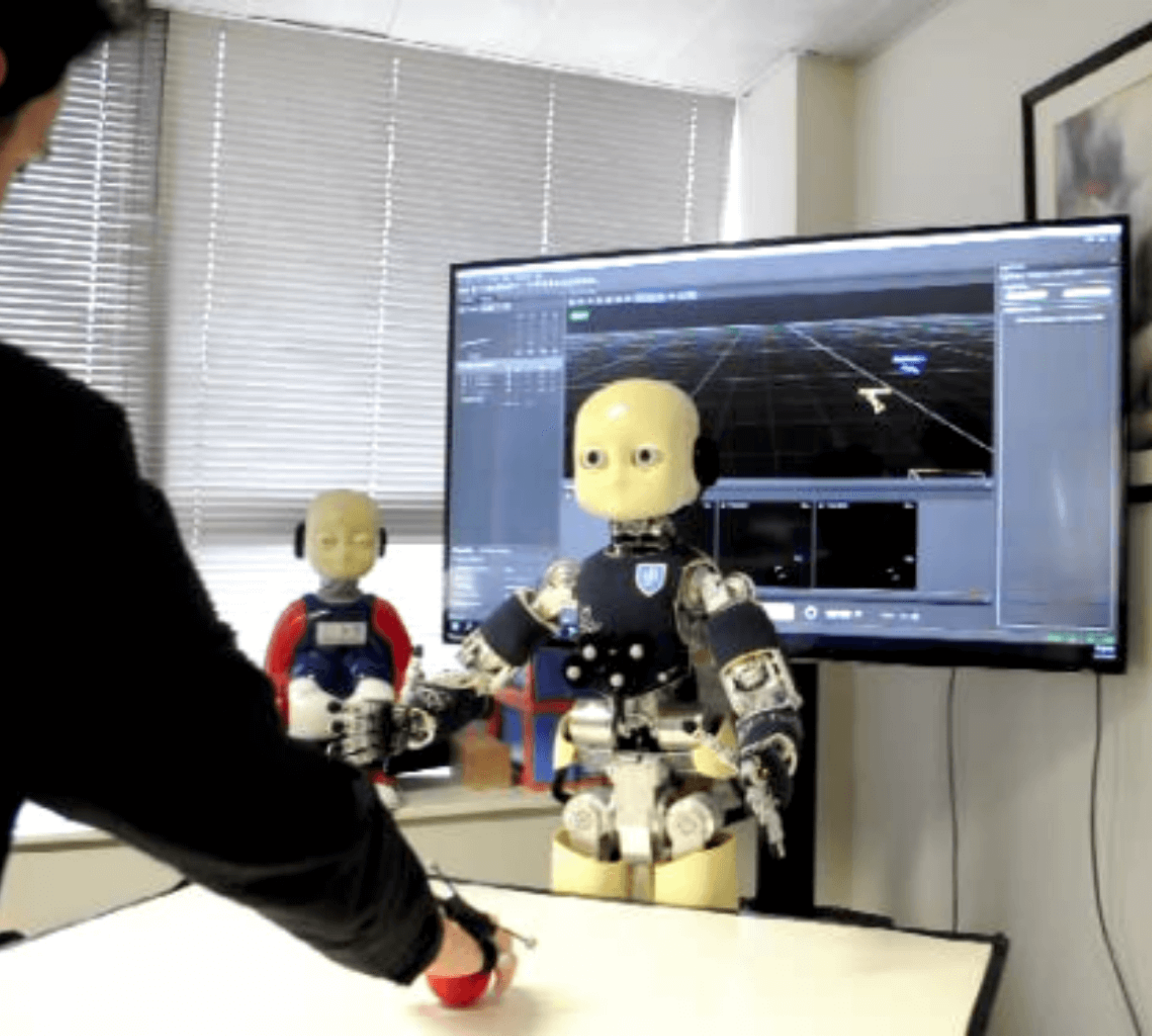}
	\includegraphics[width = 0.115\textwidth]{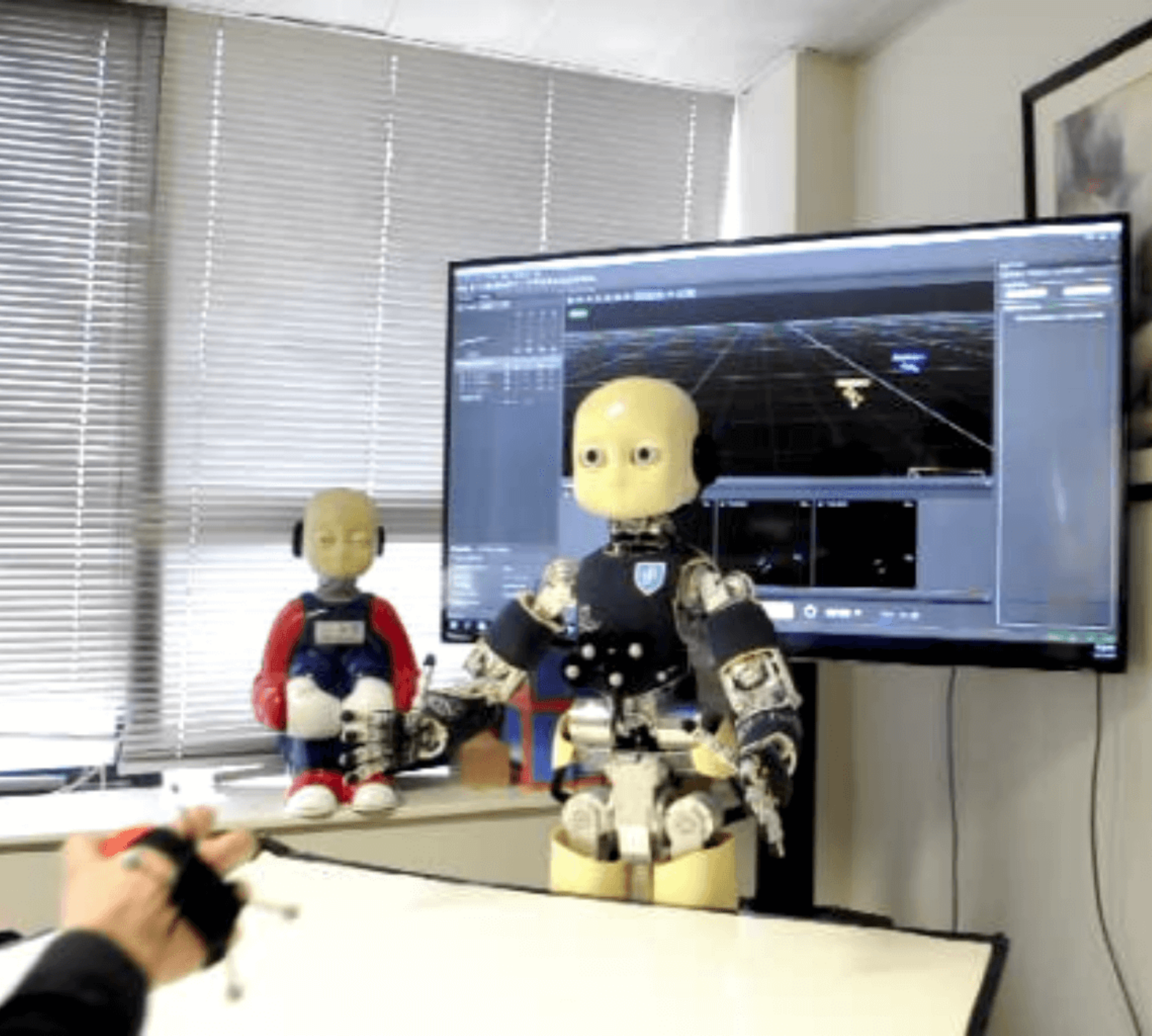}
	\includegraphics[width = 0.115\textwidth]{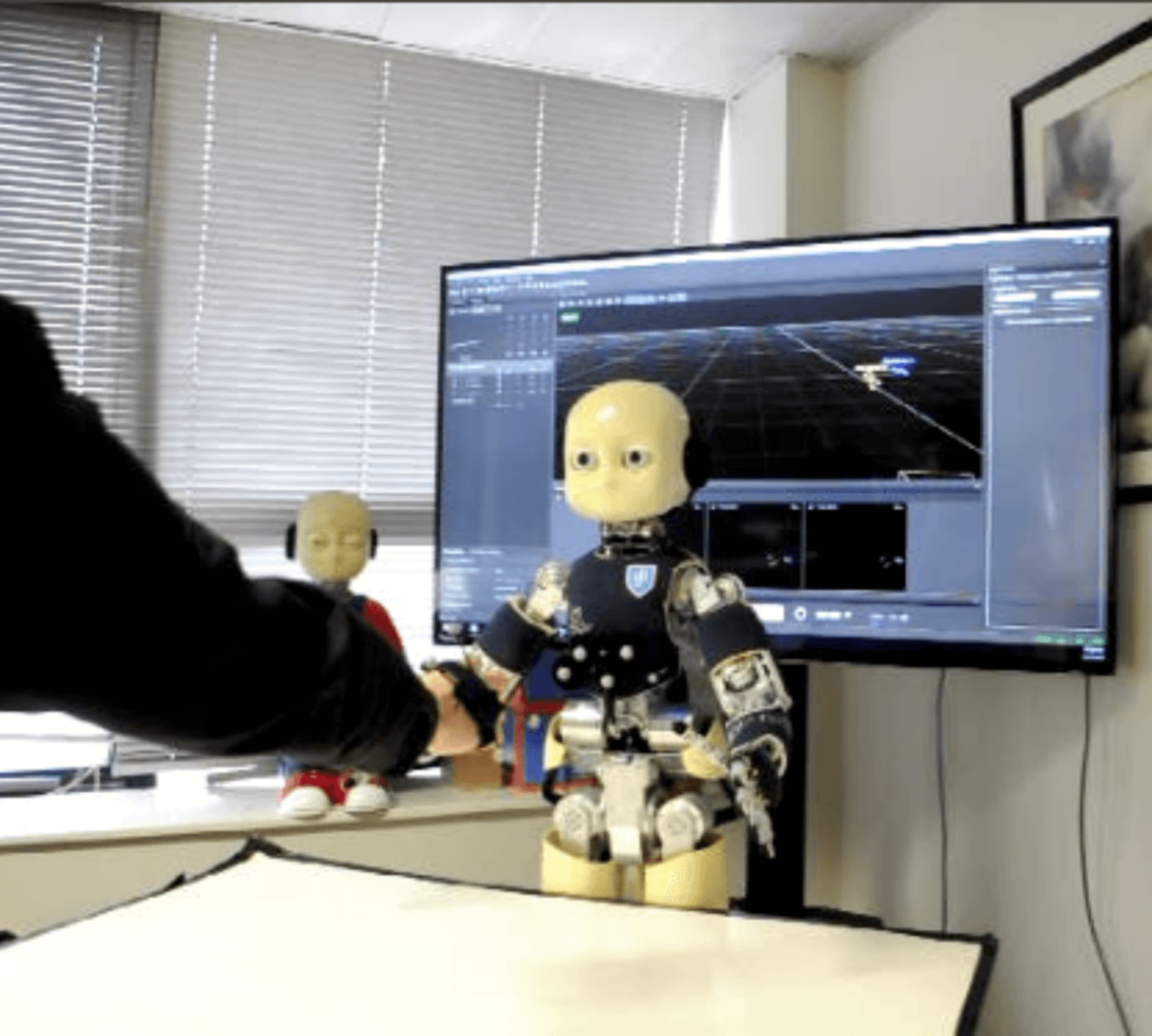}
	\includegraphics[width = 0.115\textwidth]{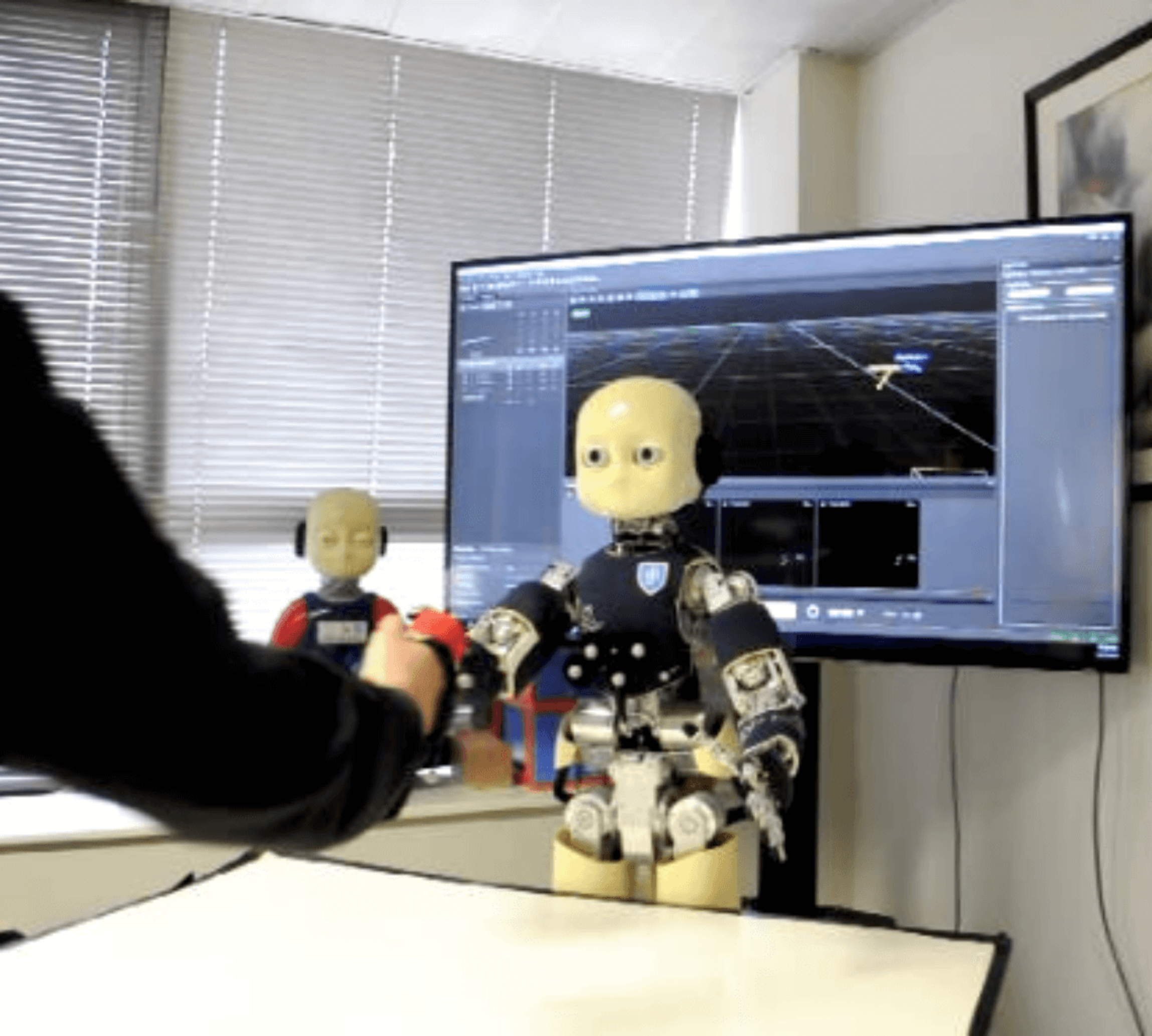}
	\includegraphics[width = 0.115\textwidth]{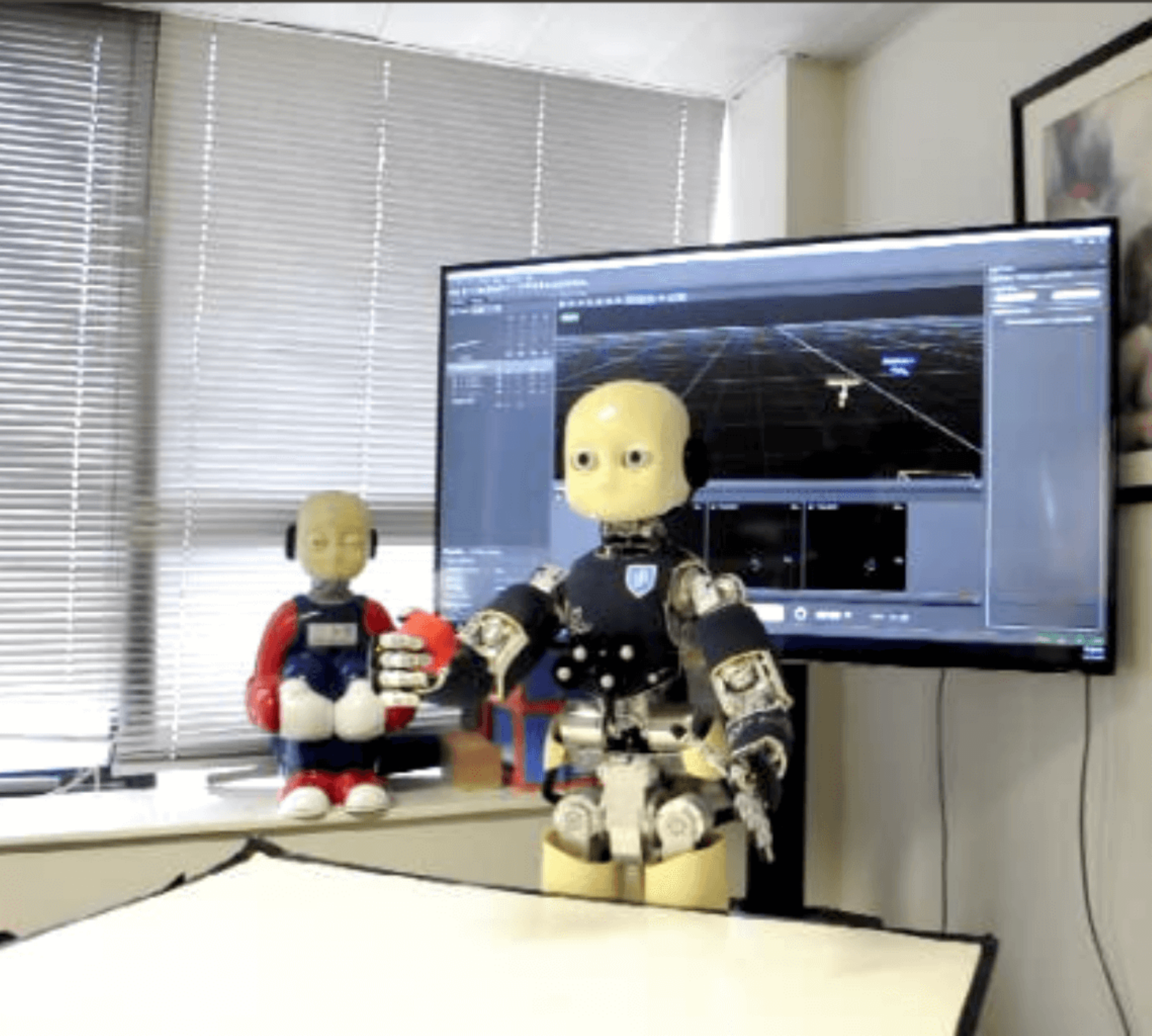}
\caption{HRI experiment involving a human handing over an object to the iCub. This experiment exemplifies the adaptability to human behaviour. The human begins by placing the object in front of the robot and, due to the coupling functions, the action is not recognized as handover, so the robot does not interact with the human. Only when the action is recognized as a handover does the robot behave to receive the object. The HRI experiments are demonstrated in the complementary video.}\label{fig:hri-exp}
\end{figure*}

The following assumptions were made during our experiments. Since our dataset contains only information regarding the wrist, we pre-determined the hand aperture and grasp orientation according to our HHI experiments. Another assumption is regardin the handover location. In the dataset the point was computed as the average point around all the experiments. For the HRI experiments, the handover location was set as the furthest point the iCub humanoid robot could reach without moving his torso. The height of the handover location was set according to our previous experiments, and the orientation of the handover was updated, in real-time, according to the orientation of the human wrist,with respect to the iCub torso. These assumptions allows us to have an adaptive handover dependent on the trained dataset. 

Our results prove that the robot performs a biologically inspired motion during the interaction with a human handing over an object. Moreover, the coupling function allows the robot to understand the intention of the human and decode the handover action. Figure \ref{fig:hri-exp} shows one of the experiments performed with the robot where the coupling comes into play. During the experiment the human performs three actions: placing an object in front of the robot, picking up the same object, handing over the object to the robot. From the understanding of the behaviour of the arm motion in handover actions, it is possible to disseminate between non-handover actions, and handover actions. The internal model of the leader in Section \ref{sec:agent} helps understand the intention of the human, and the coupling model in Section \ref{sec:couple} is responsible of adapting the motion of the robot to the observed behaviour. 
    
\section{DISCUSSION}  \label{sec:discuss}

In this paper, we presented a CDS approach for learning agent-to-agent coordination from human demonstration. Our first contribution involved modelling the coordination between the arm movements of a person handing in the object and another person receiving it. 
The modelling of this action coupling extends upon previous work \cite{duarte2018action} of generating human-like behaviour of action-in-interaction scenarios to the agent reacting to on-going interactions (follower). 
The second contribution relates to the unconscious synchronisation happening between two agents when collaborating in a shared goal. A coupling dynamical system was applied to model the intricacies between arm motions during a handover movement. Two separate coupling functions were gathered from human demonstrations and allowed for a detailed understanding of human awareness of action intention during action observation. The third contribution focused on developing controllers from the computational models for a robotic humanoid robot in order to: (i) understand from the human arm motion the intention for handing an object, and (ii) for a handover action, behave in a biologically inspired way to receive the object legible for the human. 

In the future we intend to extend this work to fulfill a complete non-verbal communication model. The integration of the motor coordination with the visual feedback would allow for a visual-motor agent-to-agent coordination which is a fare representation of how humans interact with the world and others. Another step that would be important to complement the motor coordination, is to understand the coupling of the grip aperture with the arm motion during handover actions. 

%%%%%%%%%%%%%%%%%%%%%%%%%%%%%%%%%%%%%%%%%%%%%%%%%%%%%%%%%%%%%%%%%%%%%%%%%%%%%%%%

\addtolength{\textheight}{0cm}   % This command serves to balance the column lengths

%%%%%%%%%%%%%%%%%%%%%%%%%%%%%%%%%%%%%%%%%%%%%%%%%%%%%%%%%%%%%%%%%%%%%%%%%%%%%%%%

\section*{ACKNOWLEDGMENT}

The authors wish to thank Luka Lukic for his most valuable comments and suggestions on an earlier version of the paper.

%%%%%%%%%%%%%%%%%%%%%%%%%%%%%%%%%%%%%%%%%%%%%%%%%%%%%%%%%%%%%%%%%%%%%%%%%%%%%%%%
\bibliographystyle{ieeetr}
\bibliography{IEEEabrv,mybibfile}

\begin{thebibliography}{10}

\bibitem{SEBANZ2006JointAction}
N.~Sebanz, H.~Bekkering, and G.~Knoblich, ``Joint action: bodies and minds
  moving together,'' {\em Trends in Cognitive Sciences}, vol.~10, no.~2, pp.~70
  -- 76, 2006.

\bibitem{Rozzi2015Grasping}
S.~Rozzi and G.~Coudé, ``Grasping actions and social interaction: neural bases
  and anatomical circuitry in the monkey,'' {\em Frontiers in Psychology},
  vol.~6, p.~973, 2015.

\bibitem{ibarra2018synchronization}
L.~S. Pesce~Ibarra, ``Synchronization matters for motor coordination,'' {\em
  Journal of Neurophysiology}, vol.~119, no.~3, pp.~767--770, 2018.
\newblock PMID: 28978763.

\bibitem{nowak2017functional}
A.~Nowak, R.~R. Vallacher, M.~Zochowski, and A.~Rychwalska, ``Functional
  synchronization: The emergence of coordinated activity in human systems,''
  {\em Frontiers in psychology}, vol.~8, p.~945, 2017.

\bibitem{huyi2017brain-brain}
Y.~Hu, Y.~Hu, X.~Li, Y.~Pan, and X.~Cheng, ``Brain-to-brain synchronization
  across two persons predicts mutual prosociality,'' {\em Social Cognitive and
  Affective Neuroscience}, vol.~12, no.~12, pp.~1835--1844, 2017.

\bibitem{bassetti2017action-in-interaction}
C.~Bassetti, ``Chapter 2 - social interaction in temporary gatherings: A
  sociological taxonomy of groups and crowds for computer vision
  practitioners,'' in {\em Group and Crowd Behavior for Computer Vision}
  (V.~Murino, M.~Cristani, S.~Shah, and S.~Savarese, eds.), pp.~15 -- 28,
  Academic Press, 2017.

\bibitem{rakovic2018dataset}
M.~Rakovi{\'c}, N.~Duarte, J.~Tasevski, J.~Santos-Victor, and B.~Borovac, ``A
  dataset of head and eye gaze during dyadic interaction task for modeling
  robot gaze behavior,'' in {\em MATEC Web of Conferences}, vol.~161, p.~03002,
  EDP Sciences, 2018.

\bibitem{sciutti2018humanizing}
A.~Sciutti, M.~Mara, V.~Tagliasco, and G.~Sandini, ``Humanizing human-robot
  interaction: On the importance of mutual understanding,'' {\em IEEE
  Technology and Society Magazine}, vol.~37, no.~1, pp.~22--29, 2018.

\bibitem{lukic2014learning}
L.~Lukic, J.~Santos-Victor, and A.~Billard, ``Learning robotic eye--arm--hand
  coordination from human demonstration: a coupled dynamical systems
  approach,'' {\em Biological cybernetics}, vol.~108, no.~2, pp.~223--248,
  2014.

\bibitem{duarte2018action}
N.~F. Duarte, M.~Rakovi{\'c}, J.~Tasevski, M.~I. Coco, A.~Billard, and
  J.~Santos-Victor, ``Action anticipation: Reading the intentions of humans and
  robots,'' {\em IEEE Robotics and Automation Letters}, vol.~3, pp.~4132--4139,
  Oct 2018.

\bibitem{mirrazavi2018unified}
S.~S. Mirrazavi~Salehian, N.~Figueroa, and A.~Billard, ``A unified framework
  for coordinated multi-arm motion planning,'' {\em The International Journal
  of Robotics Research}, p.~0278364918765952, 2018.

\bibitem{rakovic2018gazedialogue}
M.~Rakovi{\'c}, N.~Duarte, J.~Marques, and J.~Santos-Victor, ``Modelling the
  gaze dialogue: Non-verbal communication in human-human and human-robot
  interaction,'' {\em Paper under revision}, vol.~1, no.~1, pp.~1--12, 2018.

\bibitem{marin2009interpersonal}
L.~Marin, J.~Issartel, and T.~Chaminade, ``Interpersonal motor coordination:
  From human-human to human-robot interactions,'' vol.~10, pp.~479--504, 12
  2009.

\bibitem{duarte2018actionalignment}
N.~F. Duarte, M.~Rakovi{\'{c}}, J.~Marques, and J.~Santos-Victor, ``Action
  alignment from gaze cues in human-human and human-robot interaction,'' in
  {\em Computer Vision -- ECCV 2018 Workshops} (L.~Leal-Taix{\'e} and S.~Roth,
  eds.), (Cham), pp.~197--212, Springer International Publishing, 2019.

\bibitem{dragan2013legibility}
A.~D. Dragan, K.~C.~T. Lee, and S.~S. Srinivasa, ``Legibility and
  predictability of robot motion,'' in {\em 2013 8th ACM/IEEE International
  Conference on Human-Robot Interaction (HRI)}, pp.~301--308, March.

\bibitem{sisbot2010synthesizing}
E.~A. Sisbot, L.~F. Marin-Urias, X.~Broquere, D.~Sidobre, and R.~Alami,
  ``Synthesizing robot motions adapted to human presence,'' {\em International
  Journal of Social Robotics}, vol.~2, no.~3, pp.~329--343, 2010.

\bibitem{ding2011human}
H.~Ding, G.~Rei{\ss}ig, K.~Wijaya, D.~Bortot, K.~Bengler, and O.~Stursberg,
  ``Human arm motion modeling and long-term prediction for safe and efficient
  human-robot-interaction,'' in {\em Robotics and Automation (ICRA), 2011 IEEE
  International Conference on}, pp.~5875--5880, IEEE, 2011.

\bibitem{admoni2014deliberate}
H.~Admoni, A.~Dragan, S.~S. Srinivasa, and B.~Scassellati, ``Deliberate delays
  during robot-to-human handovers improve compliance with gaze communication,''
  in {\em Proceedings of the 2014 ACM/IEEE international conference on
  Human-robot interaction}, pp.~49--56, ACM, 2014.

\bibitem{mortl2014rhythm}
A.~M{\"o}rtl, T.~Lorenz, and S.~Hirche, ``Rhythm patterns
  interaction-synchronization behavior for human-robot joint action,'' {\em
  PloS one}, vol.~9, no.~4, p.~e95195, 2014.

\bibitem{andry2011using}
P.~Andry, A.~Blanchard, and P.~Gaussier, ``Using the rhythm of nonverbal
  human--robot interaction as a signal for learning,'' {\em IEEE Transactions
  on Autonomous Mental Development}, vol.~3, no.~1, pp.~30--42, 2011.

\bibitem{hasnain2012synchrony}
S.~K. Hasnain, G.~Mostafaoui, and P.~Gaussier, ``A synchrony-based perspective
  for partner selection and attentional mechanism in human-robot interaction,''
  {\em Paladyn}, vol.~3, no.~3, pp.~156--171, 2012.

\bibitem{khansari2011learning}
S.~M. Khansari-Zadeh and A.~Billard, ``Learning stable nonlinear dynamical
  systems with gaussian mixture models,'' {\em IEEE Transactions on Robotics},
  vol.~27, no.~5, pp.~943--957, 2011.

\bibitem{shukla2012coupled}
A.~Shukla and A.~Billard, ``Coupled dynamical system based armâhand
  grasping model for learning fast adaptation strategies,'' {\em Robotics and
  Autonomous Systems}, vol.~60, no.~3, pp.~424--440, 2012.
\newblock Autonomous Grasping.

\bibitem{mitz1991learning}
A.~R. Mitz, M.~Godschalk, and S.~P. Wise, ``Learning-dependent neuronal
  activity in the premotor cortex: activity during the acquisition of
  conditional motor associations,'' {\em Journal of Neuroscience}, vol.~11,
  no.~6, pp.~1855--1872, 1991.

\bibitem{kelso2013outline}
J.~S. Kelso, G.~Dumas, and E.~Tognoli, ``Outline of a general theory of
  behavior and brain coordination,'' {\em Neural Networks}, vol.~37,
  pp.~120--131, 2013.

\bibitem{kendon1990conducting}
A.~Kendon, {\em Conducting interaction: Patterns of behavior in focused
  encounters}, vol.~7.
\newblock CUP Archive, 1990.

\bibitem{feldman2007synchrony}
R.~Feldman, ``Parentâinfant synchrony: Biological foundations and
  developmental outcomes,'' {\em Current Directions in Psychological Science},
  vol.~16, no.~6, pp.~340--345, 2007.

\bibitem{nasrabadi2007pattern}
N.~M. Nasrabadi, ``Pattern recognition and machine learning,'' {\em Journal of
  electronic imaging}, vol.~16, no.~4, p.~049901, 2007.

\bibitem{kothe2018lab}
C.~Kothe, ``Lab streaming layer (lsl),'' {\em https://github.
  com/sccn/labstreaminglayer. Accessed on February}, 2015.

\end{thebibliography}

\end{document}